\pdfoutput=1

\documentclass[11pt]{article}

\usepackage[preprint]{acl}

\usepackage{times}
\usepackage{latexsym}

\usepackage[T1]{fontenc}

\usepackage[utf8]{inputenc}

\usepackage{microtype}

\usepackage{inconsolata}

\usepackage{graphicx}
\usepackage{adjustbox}

\newcommand{\dialemma}{\textsc{DiaLemma}}

\usepackage{amsmath}
\usepackage{booktabs}
\usepackage{arydshln}
\usepackage{makecell}
\usepackage{tabularx}
\usepackage{minted}
\usepackage[table, dvipsnames]{xcolor}
\definecolor{blanchedalmond}{rgb}{1.0, 0.92, 0.8}
\definecolor{mylightgreen}{HTML}{f6fcf9}

\definecolor{mistralsmall1}{HTML}{feffff}
\definecolor{mistralsmall2}{HTML}{ffffff}
\definecolor{mistralsmall3}{HTML}{feffff}
\definecolor{mistralsmall4}{HTML}{f5f9ff}
\definecolor{mistralsmall5}{HTML}{f2f7ff}
\definecolor{mistralsmall6}{HTML}{f9fbff}
\definecolor{mistralsmall7}{HTML}{4889f5}
\definecolor{mistralsmall8}{HTML}{4a8bf5}
\definecolor{mistralsmall9}{HTML}{4285f4}

\definecolor{mistrallarge1}{HTML}{72a4f7}
\definecolor{mistrallarge2}{HTML}{a1c3fa}
\definecolor{mistrallarge3}{HTML}{f0f5ff}
\definecolor{mistrallarge4}{HTML}{fdfeff}
\definecolor{mistrallarge5}{HTML}{e0ebfe}
\definecolor{mistrallarge6}{HTML}{fcfdff}
\definecolor{mistrallarge7}{HTML}{cddffd}
\definecolor{mistrallarge8}{HTML}{bbd3fc}
\definecolor{mistrallarge9}{HTML}{5491f6}

\definecolor{llama3.11}{HTML}{f2f7ff}
\definecolor{llama3.12}{HTML}{f8fbff}
\definecolor{llama3.13}{HTML}{fefeff}
\definecolor{llama3.14}{HTML}{e4edfe}
\definecolor{llama3.15}{HTML}{e1ecfe}
\definecolor{llama3.16}{HTML}{fafcff}
\definecolor{llama3.17}{HTML}{649bf6}
\definecolor{llama3.18}{HTML}{6099f6}
\definecolor{llama3.19}{HTML}{4285f4}

\definecolor{llama3.31}{HTML}{5692f6}
\definecolor{llama3.32}{HTML}{679df7}
\definecolor{llama3.33}{HTML}{cfe0fd}
\definecolor{llama3.34}{HTML}{ffffff}
\definecolor{llama3.35}{HTML}{fcfdff}
\definecolor{llama3.36}{HTML}{feffff}
\definecolor{llama3.37}{HTML}{e4eefe}
\definecolor{llama3.38}{HTML}{d7e5fd}
\definecolor{llama3.39}{HTML}{6ca1f7}

\definecolor{llama41}{HTML}{f4f8ff}
\definecolor{llama42}{HTML}{feffff}
\definecolor{llama43}{HTML}{ffffff}
\definecolor{llama44}{HTML}{699ef7}
\definecolor{llama45}{HTML}{5c96f6}
\definecolor{llama46}{HTML}{b9d2fb}
\definecolor{llama47}{HTML}{fbfcff}
\definecolor{llama48}{HTML}{fdfeff}
\definecolor{llama49}{HTML}{e4eefe}
\definecolor{llama410}{HTML}{e1ecfe}
\definecolor{llama411}{HTML}{e2ecfe}
\definecolor{llama412}{HTML}{9dc0fa}

\definecolor{ayasmall1}{HTML}{b3cefb}
\definecolor{ayasmall2}{HTML}{b5d0fb}
\definecolor{ayasmall3}{HTML}{f6f9ff}
\definecolor{ayasmall4}{HTML}{ffffff}
\definecolor{ayasmall5}{HTML}{ffffff}
\definecolor{ayasmall6}{HTML}{ffffff}
\definecolor{ayasmall7}{HTML}{87b2f9}
\definecolor{ayasmall8}{HTML}{86b1f8}
\definecolor{ayasmall9}{HTML}{4487f5}

\definecolor{ayalarge1}{HTML}{a3c4fa}
\definecolor{ayalarge2}{HTML}{bcd4fc}
\definecolor{ayalarge3}{HTML}{f2f7ff}
\definecolor{ayalarge4}{HTML}{f1f6ff}
\definecolor{ayalarge5}{HTML}{e6effe}
\definecolor{ayalarge6}{HTML}{fbfdff}
\definecolor{ayalarge7}{HTML}{a9c7fa}
\definecolor{ayalarge8}{HTML}{9abefa}
\definecolor{ayalarge9}{HTML}{4d8df5}
\definecolor{ayalarge10}{HTML}{fcfdff}
\definecolor{ayalarge11}{HTML}{fdfeff}
\definecolor{ayalarge12}{HTML}{ffffff}

\definecolor{gemmasmall1}{HTML}{75a6f7}
\definecolor{gemmasmall2}{HTML}{90b7f9}
\definecolor{gemmasmall3}{HTML}{e0ebfe}
\definecolor{gemmasmall4}{HTML}{f0f5ff}
\definecolor{gemmasmall5}{HTML}{d5e4fd}
\definecolor{gemmasmall6}{HTML}{f6faff}
\definecolor{gemmasmall7}{HTML}{b4cffb}
\definecolor{gemmasmall8}{HTML}{b4cffb}
\definecolor{gemmasmall9}{HTML}{4285f4}

\definecolor{gemmalarge1}{HTML}{4c8bf5}
\definecolor{gemmalarge2}{HTML}{8fb7f9}
\definecolor{gemmalarge3}{HTML}{9bbffa}
\definecolor{gemmalarge4}{HTML}{e6effe}
\definecolor{gemmalarge5}{HTML}{9dc0fa}
\definecolor{gemmalarge6}{HTML}{ebf2fe}
\definecolor{gemmalarge7}{HTML}{e7f0fe}
\definecolor{gemmalarge8}{HTML}{edf4fe}
\definecolor{gemmalarge9}{HTML}{93b9f9}

\usepackage{fontawesome5}

\newcommand{\mistralsmall}{\texttt{Mistral-7b}}
\newcommand{\mistrallarge}{\texttt{Mistral-123b}}
\newcommand{\llamathreesmall}{\texttt{Llama3.1-8b}}
\newcommand{\llamathreelarge}{\texttt{Llama3.3-70b}}
\newcommand{\llamafour}{\texttt{Llama4-17b}}
\newcommand{\ayasmall}{\texttt{Aya-expanse-8b}}
\newcommand{\ayalarge}{\texttt{Aya-expanse-32b}}
\newcommand{\gemmasmall}{\texttt{Gemma3-12b}}
\newcommand{\gemmalarge}{\texttt{Gemma3-27b}}

\newcommand{\iferr}{IF Error}
\usepackage{multirow}

\title{Make Every Letter Count: Building Dialect Variation Dictionaries \\ from Monolingual Corpora}

\author{Robert Litschko\textsuperscript{1,2} \quad
        Verena Blaschke\textsuperscript{1,2} \quad
        Diana Burkhardt\textsuperscript{1} \\ 
        \textbf{Barbara Plank\textsuperscript{1,2}} \quad
       \textbf{Diego Frassinelli\textsuperscript{1}} \quad\\
  \textsuperscript{1} MaiNLP, Ludwig Maximilian University of Munich, Germany \\
  \textsuperscript{2} Munich Center for Machine Learning (MCML), Munich, Germany \\
{\tt robert.litschko@lmu.de} 
}

\begin{document}
\maketitle
\begin{abstract}
Dialects exhibit a substantial degree of variation due to the lack of a standard orthography. At the same time, the ability of Large Language Models (LLMs) to process dialects remains largely understudied. To address this gap, we use Bavarian as a case study and investigate the lexical dialect understanding capability of LLMs by examining how well they recognize and translate dialectal terms across different parts-of-speech. To this end, we introduce \dialemma{}, a novel annotation framework for creating dialect variation dictionaries from monolingual data only, and use it to compile a ground truth dataset consisting of 100K human-annotated German-Bavarian word pairs. We evaluate how well nine state-of-the-art LLMs can judge Bavarian terms as dialect translations, inflected variants, or unrelated forms of a given German lemma. Our results show that LLMs perform best on nouns and lexically similar word pairs, and struggle most in distinguishing between direct translations and inflected variants. Interestingly, providing additional context in the form of example usages improves the translation performance, but reduces their ability to recognize dialect variants. This study highlights the limitations of LLMs in dealing with orthographic dialect variation and emphasizes the need for future work on adapting LLMs to dialects.\footnote{\url{https://github.com/mainlp/dialemma}}
\end{abstract}

\section{Introduction}

Although most languages have standardized orthographies, this assumption does not hold for many languages and language varieties \cite{millour-fort-2020-text}.
One prominent example are non-standard dialects, which have become increasingly popular in natural language processing (NLP) research \cite{faisal-etal-2024-dialectbench, Joshi2025SurveyDialects}.

Dialects exhibit a large degree of spelling variation in written data, owing both to individual pronunciation differences and spelling preferences \cite{just2025corpus}.
Such variation is commonly reflected in dialect NLP datasets
\cite[cf.][]{blaschke-etal-2023-survey, ramponi-2024-language}.
However, current NLP methods -- which typically are trained on large amounts of standard-language data -- are badly equipped to deal with such spelling variation, even if the word forms are similar to the standard.
Issues range from subword tokenization \cite{blaschke-etal-2023-manipulating, srivastava-chiang-2025-calling}, to evaluating generative tasks \cite{aepli-etal-2023-benchmark}, to retrieving information without being able to match word spellings \cite{litschko-etal-2025-cross}.
\begin{table}[t]
\centering
\adjustbox{max width=\columnwidth}{%
\begin{tabular}{@{}llr@{}}
\multicolumn{3}{@{}c@{}}{{\textbf{1.\ Judging Translation Pairs}}\vspace{0.1cm}}\\
\toprule
{German} & {Bavarian} & \llap{Translation}?\\
\midrule
\textit{zweisprachig} & \textit{zwaasprochig} & yes \\
(``bilingual'') & \textit{zwaspråchig} & yes \\
& \textit{zwoasprachign} & infl.\\
& \textit{dreisprochige} {(``trilingual'')} & no \\
& ... & ... \\ 
\textit{dazwischen} & \textit{dozwischn} & yes \\
(``in between'') & \textit{dawischn} {(``to catch'')} & no \\
& \textit{daktischen} {(``tactical'')} & no \\
& ... & ... \\ 
\bottomrule
\end{tabular}
}
\\[6pt]
\adjustbox{max width=\columnwidth}{%
\begin{tabular}{@{}lll@{}}
\multicolumn{3}{@{}c@{}}{{\textbf{2.\ Dialect-to-Standard Translation}}\vspace{0.1cm}} \\
\toprule
Bavarian & & German lemma \\
\midrule
\textit{zwaasprochig} & $\rightarrow$ & \textit{zweisprachig} \\
\textit{zwaspråchig} & $\rightarrow$ & \textit{zweisprachig} \\
\textit{dozwischn} & $\rightarrow$ & \textit{dazwischen} \\
... & $\rightarrow$ & ...\\
\bottomrule
\end{tabular}
}
\caption{We annotate whether Bavarian dialect words with high string similarity to Standard German lemmas are (direct or inflected) translations~(1). We test whether LLMs are able to do the same, and whether they can translate dialectal words into their standard-language counterparts~(2).}
\label{tab:example}
\end{table}
Therefore, it is important to build NLP systems that are either robust to  variation or good at normalization. In order to gauge how good current or future systems are at this, we need evaluation datasets. However, prior works on collecting datasets of real-life dialectal spelling variation have resulted in relatively small datasets~(Section~\ref{sec:related-work}).

To allow large-scale analyses of the impact of dialectal spelling variation on NLP tasks, we conduct a case study on the Wikipedia edition in Bavarian, a dialect group related to German. This dialect wiki is comparatively large and has been used in prior work on dialect NLP \cite{artemova-plank-2023-low, peng-etal-2024-sebastian, blaschke-etal-2024-maibaam, litschko-etal-2025-cross}, and is included in pre-training datasets for, e.g., mBERT \cite{devlin-etal-2019-bert}. It covers the entire Bavarian-speaking area (regardless of subdialect), and explicitly encourages contributors to spell words as they find appropriate.\footnote{\href{https://bar.wikipedia.org/wiki/Wikipedia:Wia_schreib_i_a_guads_Boarisch\%3F}{\texttt{bar.wikipedia.org/\allowbreak{}wiki/\allowbreak{}Wikipedia:\allowbreak{}Wia\_\allowbreak{}schreib\_\allowbreak{}i\_\allowbreak{}a\_\allowbreak{}guads\_\allowbreak{}Boarisch?}} ``How do I write good Bavarian?''.}

Our approach only requires access to one monolingual corpus per language variety. Taking advantage of the fact that related standard and dialect varieties have a high number of cognates with similar word forms,\footnote{We acknowledge that there are other kinds of differences between standard and dialect varieties, e.g., lexical differences due to regional word choices. In this study, we focus on \textit{spelling differences between cognate words} as the spelling variation is a specific challenge for NLP systems.} we automatically extract German--Bavarian word pairs with high string similarities and annotate whether they are (direct or inflected) translations of each other (Table~\ref{tab:example}). We use our dataset to systematically analyze how well recent large language models (LLMs) can perform such annotations, and how good they are at translating the Bavarian terms into their German counterparts. Our study makes the following contributions:

\begin{itemize}
    \item We introduce a \textbf{novel annotation framework} for creating dialect variation dictionaries from monolingual data, without requiring parallel corpora~(Section~\ref{sec:dataset-pipeline}).
    
    \item For 10k German words, we extract and manually annotate similarly spelled Bavarian words, resulting in \textbf{11k direct German--Bavarian translation pairs} and \textbf{7k pairs with inflection differences}~(Section~\ref{sec:dataset-stats}).
    
    \item We \textbf{evaluate nine state-of-the-art LLMs} on two tasks: \textbf{judging} whether a given word pair is a translation~(Section~\ref{sec:results-recognition}), and \textbf{translating} Bavarian words into their German counterparts~(Section~\ref{sec:results-translation}). We show that, overall, larger models perform better than their smaller versions, but they still struggle to distinguish between translations and inflected variants (Section~\ref{sec:results-additional}).
    
    \item We further show that (i) including additional context in the form of usage examples, on average, improves the model's ability to translate and hurts the performance in judging translation candidates, (ii) using German prompts leads to worse results, and (iii) the Levenshtein distance between word pairs crucially impacts the model performance (Section~\ref{sec:results-additional}). 
\end{itemize}

\begin{figure*}[t!]
    \centering
    \includegraphics[width=0.95\linewidth]{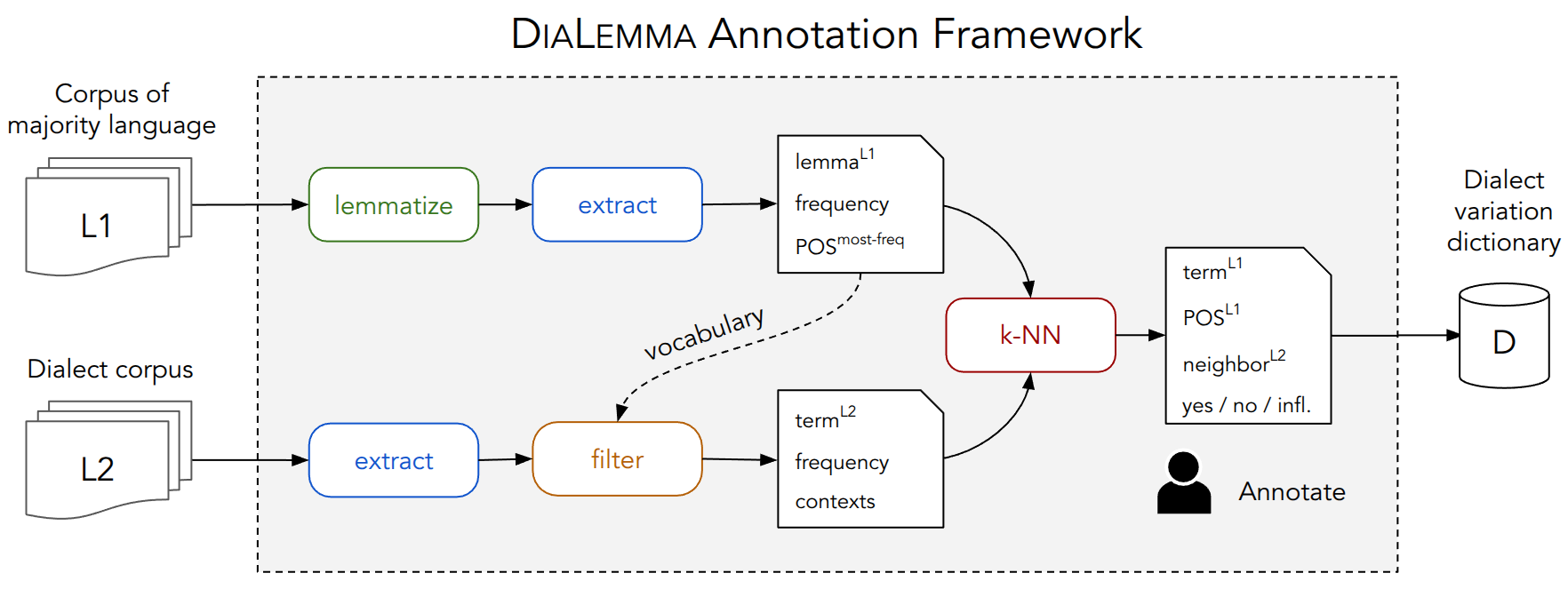}
    \caption{\dialemma{} annotation framework. Step~1: Lemmatize the corpus of the majority language. Step~2: Extract the  vocabularies from both the corpora of the majority language and dialect, respectively. Step~3: Filter the lemmas of the majority language out of the dialect vocabulary. Step~4: Find for each lemma the lexically most similar dialect terms using the Levenshtein distance. Step~5: Annotate if dialect terms are translations (`yes'), dialectal inflected forms (`inflected'), or neither (`no').}
    \label{fig:dialemma}
\end{figure*}

\section{Related Work}
\label{sec:related-work}

\paragraph{Inducing dialect and low-resource language dictionaries} \citet{artemova-plank-2023-low} induce dialect dictionaries from parallel articles in German and dialectal wikis, extracting $\sim$800 word pairs per language pair, and manually verifying them. 
Similarly, \citet{haddow-etal-2013-corpus} extract German--Bavarian word lists from parallel and near-parallel sentences.
Unlike these works, our approach does not rely on parallel data.

\citet{burghardt-etal-2016-creating} and \citet{burghardt2023bairisch} crowdsourced German translations for 259 Bavarian words, while \citet{millour-fort-2019-unsupervised} collected spelling variants for 145 Alsatian words, resulting in about seven variants per word. \citet{schmidt-etal-2020-swiss} manually translated a German word list into Swiss German dialects. Compared to these efforts, we do not require large-scale manual translations.

\citet{litschko-etal-2025-cross} automatically induced translations for dialectal terms in seven German dialects and regional languages based on Wikipedia article titles and links, also collecting some spelling (and, rarely, lexical) variations. Unlike this approach, we are not limited to wiki article topics. \citet{ylonen-2022-wiktextract} uses Wiktionary as a structured data source for extracting word-level translations (with inflection information). However, dialects are not thoroughly covered in Wiktionary.

The idea behind finding dialect/standard translation pairs is similar to work on identifying cognate pairs in more distantly related languages, for which complex statistical alignment methods have been developed \cite[\textit{inter alia}]{inkpen-etal-2005-automatic,haghighi-etal-2008-learning,wettig-etal-2011-mdl,kontonatsios-etal-2014-combining}. We are however interested in finding word pairs in closely related varieties, and we allow many-to-one dialect-to-standard mappings to account for spelling variations. \citet{li-etal-2023-bilingual} fine-tune LLMs to predict word-level translations between pairs of languages, showing that models are sensitive to prompt choices and perform poorly on low-resource languages. 

\paragraph{Dialect-to-standard translation and normalization} Translating dialectal words and texts into the corresponding standard language is a well-established task within dialect NLP \cite{zampieri2020survey}. Similar efforts have focused on normalizing historical spellings \cite{bollmann-2019-large}, and research on normalizing social media data has also included normalizing phonetic spellings or slang terms into the corresponding standard language forms  \cite{van-der-goot-etal-2018-taxonomy}. Popular strategies for normalizing dialect spellings on a word or sentence level use character-based machine translation \citep{honnet-etal-2018-machine, kuparinen-etal-2023-dialect, scherrer-2023-character}. More recent works have also treated sentence-level dialect-to-standard translation like a common translation task, using encoder-decoder transformer architectures with subword tokens \cite{kuparinen-etal-2023-dialect, kresic2024normalizing, her-kruschwitz-2024-investigating}.

Our focus lies on instruction-tuned LLMs, as they might normalize data in an unsupervised way. \citet{alam-anastasopoulos-2025-large} prompt instruction-tuned LLMs for sentence-level dialect normalization, observing poor zero-shot performances. We study LLMs on the word level.

\section{Method}
\label{sec:dataset}

In this section, we first introduce \textsc{DiaLemma}, our annotation framework for obtaining dialect variation dictionaries~(Section~\ref{sec:dataset-pipeline}). We then introduce our dataset~(Section~\ref{sec:dataset-stats}), which we obtained by applying \textsc{DiaLemma} to the Bavarian Wikipedia. 
Based on this, we introduce two word-level dialect tasks (Section~\ref{sec:dataset-tasks}).

\subsection{\dialemma{} Annotation Framework}
\label{sec:dataset-pipeline}

Figure~\ref{fig:dialemma} shows our annotation framework for obtaining dialect variation dictionaries from monolingual corpora. We require two monolingual corpora, one in a standard language $L_1$ (e.g., German), and one in a related non-standardized dialect $L_2$ (e.g., Bavarian).
The goal is to match spelling variations in $L_2$ (\textit{zwaa\-spro\-chig, zwas\-prå\-chig, zwoa\-spro\-chig, zwa\-spro\-chig}; ``bilingual'') to their normalized counterpart in $L_1$ (\textit{zwei\-spra\-chig}). We additionally match inflected $L_2$ word forms (like \textit{zwoa\-spro\-chign})\footnote{The suffix \textit{-n} marks \textsc{def.pl}, \textsc{dat}, or \textsc{masc.acc}.} to the corresponding $L_1$ lemmas. \dialemma{} frames the extraction of dialect variation dictionaries as a one-to-many matching problem between the two vocabularies, extracted from monolingual corpora, where judgments about word matches are done by native speakers. In Section~\ref{sec:results}, we investigate to what extent LLMs can be used in place of human judges.

In principle, each dialect term could be a match for any term in the majority language. However, manually comparing all possible pairs is intractable for large vocabularies.  To reduce the search space, \dialemma{} first applies part-of-speech (POS) tagging and lemmatization on the $L_1$ corpus (we use spaCy, \citealp{spacy2}, for this).\footnote{\href{https://spacy.io/}{https://spacy.io/}, v3, \textit{de\_core\_news\_lg} model} We then extract the vocabulary of $L_1$ consisting of unique lemmas in $\mathcal{V}^{L_1}$. For each lemma, we extract also its most frequent POS tag ($\text{POS}^\text{max}$) as additional information for annotators. In practice, we limit the vocabulary to the most frequent $n$ standard-language terms. Since there are no morphosyntactic analyzers for most dialects, we directly extract the list of \textit{all} unique terms found in $L_2$. Importantly, as a result of dialectal spelling variations, many dialect terms have a low term frequency, which is why we do not limit $\mathcal{V}^{L_2}$ by frequency. To further reduce the search space, we filter out tokens that are shared between $\mathcal{V}^{L_1}$ and $\mathcal{V}^{L_2}$.

In the next step of \dialemma{}, the $k$ nearest neighbors for each lemma are extracted based on their \citet{levenshtein-1966-binary} distance (LD). If more than $k$ dialect terms have the same LD, we select $k$ of them at random. As a result, we obtain for each lemma its $\text{POS}^\text{max}$, a set of $k$ possible dialect variations (candidates) and term frequencies. To provide annotators with further context, we include up to $c$ examples in which the dialect term is used. 

The final step involves native speakers, who compare lemmas in the majority language to matched dialect terms, and annotate whether the dialect term corresponds to a dialect translation (`yes'), an inflected dialect translation (`inflected'), or neither of the two (`no') (see Table~\ref{tab:example}). For the `inflected' class, we refer to forms inflected in ways distinct from the lemma form (e.g., infinitives are considered exact matches rather than inflected variants). We distinguish between `yes' and `inflected' so that our work can be used to induce bilingual dictionaries for lemmas and inflected translations (since forms with different inflectional morphology do not correspond to direct translations). We lemmatize the German words since dictionaries usually are on the lemma level, and since this reduces the search space for finding matches between German and Bavarian words, allowing us to annotate a greater number of (entirely) different words.

All instances annotated as `yes' or `inflected' are included in the final dialect variation dictionary. This allows us to analyze the performance of LLMs with respect to different POS classes, and between dialectal translations and inflections. 

\begin{table}[t!]
\centering

\begin{tabular}{l r r r} \toprule
 & Yes & Inflected & No \\ \midrule 
Noun & 6,720 & 2,670 & 28,480 \\
Adjective & 1,358 & 3,066 & 5,496 \\
Adverb & 1,157 & --- & 2,783 \\
Verb & 934 & 1,182 & 6,214 \\ 
Proper Noun & 574 & 86 & 34,430 \\ \cdashline{1-4}[.4pt/1pt]\noalign{\vskip 0.5ex} 
Total & 11,044 & 7,070 & 81,586 \\ \midrule
Noun & 6,564 & --- & --- \\
Adjective & 1,325 & --- & --- \\
Adverb & 1,126 & --- & --- \\
Verb & 916 & --- & --- \\
Proper Noun & 556 & --- & --- \\ \cdashline{1-4}[.4pt/1pt]\noalign{\vskip 0.5ex} 
Total & 10,775 & --- & --- \\ \bottomrule
\end{tabular}
\caption{Label distributions for the test set of the judgment task (top half) and translation task (bottom half).
We show the numbers for the entire test split as well as for the five most frequent parts of speech in the test set. Statistics for all POS classes are shown in Table~\ref{tab:full_statistics}.
}
\label{tab:statistics}
\end{table}

\subsection{Dataset}
\label{sec:dataset-stats}

We use \dialemma{} on monolingual corpora we extract from the Bavarian and German Wikipedias (\href{https://creativecommons.org/licenses/by-sa/4.0/deed.en}{CC BY-SA 4.0}). In total, we collect for each of 10k German lemmas the $k=10$ nearest neighbors, each with $c=3$ local contexts, consisting of fifty characters before and after a dialect term within a sentence. Note that local contexts are not used in the extraction process. We extract it to provide additional information for human annotators and to investigate whether it improves the performance of LLMs in recognizing and translating dialect variants. This constitutes a total number of 100k Bavarian--German word pairs, out of which we hold out 300 instances as a development set for prompt selection. All instances are manually annotated by two in-house undergraduate and one postgraduate student, who are native speakers of Bavarian and German. The annotators show a high consistency in their judgments, with a Fleiss' kappa \citep{fleiss1971measuring} score of 0.86. 

\begin{figure}
    \centering
    \includegraphics[width=0.85\linewidth]{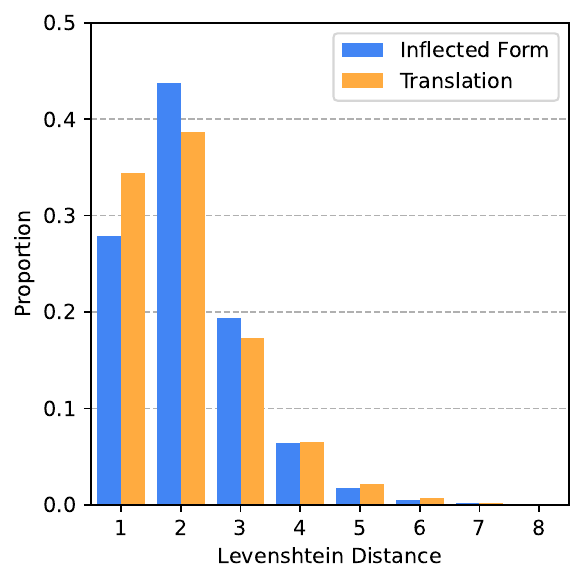}
    \caption{Distribution of word pairs by Levenshtein distance. The bars indicate the proportion of extracted translations (i.e., exact matches; in orange) versus inflected forms (in blue) at each distance value.}
    \label{fig:LDpairs}
\end{figure}

As shown in Table~\ref{tab:statistics}, our annotated dataset contains Bavarian translations for $11,044$ German lemmas, and inflected Bavarian translations for $7,070$ lemmas (test set). German lemmas for which we found direct Bavarian translations are mapped to $2.61\pm1.88$ spelling variations on average. Lemmas for which we found inflected counterparts are mapped to $2.57\pm1.90$ inflected Bavarian forms on average. Bavarian translations have an average Levenshtein distance of $2.07\pm1.07$ from their German counterparts. Inflected Bavarian terms have a Levenshtein distance of $2.13\pm1.01$. Word pairs that are \textit{not} translations of each other (`inflected' or `no') have an average Levenshtein distance of $2.72\pm1.18$. Figure~\ref{fig:LDpairs} presents the percentage of word pairs for each Levenshtein distance; the bars distinguish between Bavarian translations (in orange) and inflected forms (in blue).

\subsection{NLP Tasks: Can LLMs Understand Spelling Variation?}
\label{sec:dataset-tasks}

We now describe the two benchmarks that we build to systematically evaluate the orthographic dialect understanding capabilities of LLMs. The focus of our work lies on Bavarian and German. 

\paragraph{Judging translation candidates} We test whether LLMs can step in the role of annotators and judge whether Bavarian candidate words correspond to a given German lemma (Figure~\ref{fig:dialemma}). LLMs are presented with a word pair and need to classify their relationship into one of the three possible classes `yes', `inflected', and `no'. 
Since the class distribution is heavily skewed towards the `no' class, we choose the macro-averaged F1 score as our evaluation metric, treating all classes equally.
In Table~\ref{tab:statistics} (top), we show the label distribution for the test split of the annotated dataset. 
The word pairs included in this task have an average Levenshtein distance of $2.60\pm1.18$.
We reserve a development set for LLM prompt selection~(Section~\ref{sec:method-prompts}).
 
\paragraph{Dialect-to-standard translation} We also test the ability of LLMs translate Bavarian terms to German. For this, we use all instances with the label `yes'. Instances belonging to the other two classes  cannot be used due to the lack of reference translations. We report the word-level translation accuracy and limit the output of LLMs to at most 20 tokens. Instances where LLMs output more than one word are considered instruction-following errors.

\section{Experimental Setup}
\label{sec:experimental-setup}

\subsection{Language Models}
\label{sec:method-llms}
We conduct our experiments using nine state-of-the-art open-source large language models. 
We experiment with different model families and model sizes, including both smaller and larger variants for each family. 
Specifically, we use \ayasmall{} and \ayalarge{} \citep{dang2024aya}; \gemmasmall{} and \gemmalarge{} \citep{team2025gemma}; \llamathreesmall{}, \llamathreelarge{}, and \llamafour{} \citep{Grattafiori2024}; as well as \mistralsmall{} and \mistrallarge{} \citep{jiang2023mistral7b}. 
For all LLMs, we use the instruction-tuned version and greedy decoding (temperature\,=\,0).

\subsection{Prompt Selection}
\label{sec:method-prompts}

LLMs are known to generate different outputs under minor prompt modifications. For this reason, we develop a pool of, respectively, 22 and 13 English-language prompts for the judgment and translation task (Appendix~\ref{appendix:prompts}). We follow \citet{li-etal-2023-bilingual} and use a development set to select the best-performing prompt for each LLM. For the judgment and translation task, the development set contains 300  instances, respectively. Our main results are based on these prompts, however we carry out two additional experiments to determine the effect of the prompt selection:

\textbf{Prompts with context.} We extend the best prompts to also contain a usage example of the dialect word (see Tables~\ref{tab:recognition_w_context_prompts} and \ref{tab:translation_w_context_prompts} in Appendix~\ref{appendix:prompts}). Here, the contexts are identical to those used in the annotation process and correspond to the three first example usages found in the Wikipedia dump.

\setlength{\tabcolsep}{5pt}

\begin{table*}[t!]
\centering
\begin{tabular}{lccccc|cc}
\toprule
\textbf{Model} & \textbf{Noun} & \textbf{Adjective} & \textbf{Adverb} & \textbf{Verb} & \textbf{Proper Noun} & \textbf{Overall} & \textbf{IF Error} \\ 
\midrule
\texttt{Random} & 0.268 & 0.309 & 0.257 & 0.280 & 0.174 & 0.250 & --\\ 
\texttt{Levenshtein} & 0.355 & 0.307 & 0.331 & 0.307 & 0.202 & 0.284 & -- \\ 
\texttt{Majority Label} (`no') & 0.286 & 0.238 & 0.271 & 0.285 & 0.330 & 0.300 & -- \\ 
\texttt{Logistic Regression} & 0.351 & 0.310 & 0.270 & 0.344 & 0.312 & 0.364 & -- \\
\midrule 
\mistralsmall & 0.292 & 0.270 & 0.278 & 0.335 & 0.330 & 0.332 & 0.001 \\
\mistrallarge & \textbf{0.561} & \textbf{0.488} & \textbf{0.484} & \textbf{0.517} & \textbf{0.473} & \textbf{0.567} & 0.026 \\  \cdashline{1-8}[.4pt/1pt]\noalign{\vskip 0.5ex}
\llamathreesmall & 0.398 & 0.305 & 0.300 & 0.351 & 0.383 & 0.399 & 0.003 \\
\llamathreelarge & 0.444 & 0.369 & 0.379 & 0.340 & 0.359 & 0.428 & 0.000 \\
\llamafour & 0.354 & 0.394 & 0.237 & 0.338 & 0.309 & 0.368 & 0.438 \rlap{(!)} \\ \cdashline{1-8}[.4pt/1pt]\noalign{\vskip 0.5ex}
\ayasmall & 0.447 & 0.332 & 0.392 & 0.390 & 0.436 & 0.436 & 0.000 \\
\ayalarge & 0.500 & 0.411 & 0.442 & 0.469 & 0.414 & 0.499 & 0.005 \\ \cdashline{1-8}[.4pt/1pt]\noalign{\vskip 0.5ex}
\gemmasmall & 0.492 & 0.444 & 0.441 & 0.430 & 0.390 & 0.496 & 0.000 \\
\gemmalarge & 0.453 & 0.431 & 0.288 & 0.323 & 0.261 & 0.418 & 0.001 \\
\midrule
\textit{average} & 0.438 & 0.384 & 0.382 & 0.388 & 0.373 & - & - \\
\bottomrule
\end{tabular}
\caption{\textbf{Translation candidate judgments} -- Macro F1 scores for each model across the five most frequent POS tags. The final two columns report the average Macro F1 across 15 POS (See Table~\ref{tab:full_recognition}) and the overall percentage of outputs where LLMs failed-to-follow prompt instructions (\iferr{}). 
Boldface font indicates the \textbf{highest scores}.} 
\label{tab:RecognitionSelected}
\end{table*}

\textbf{German-language prompts.} Our original list of prompts is in English, since this is the primary language of the models we use. To determine whether prompts in the standard language of the dataset work better, we also evaluate the performance of the best-performing prompts, after translating them into German (see Tables~\ref{tab:recognition_german_prompts} and \ref{tab:translation_german_prompts} in Appendix~\ref{appendix:prompts}).

\subsection{Baselines} To assess the ability of LLMs to judge translation candidates, we compare their performance against several baselines. We include a random baseline (\texttt{Random}), a heuristic based on the Levenshtein distance (LD), where instances with a distance of two or less are classified as `yes' and otherwise as `no', and a majority label baseline (\texttt{Majority Label}) that always predicts the majority label `no'. Additionally, we train a logistic regression model on the development set, and use the Levenshtein distance and Jaccard similarity between character bi- and trigrams of the German lemma and Bavarian candidate as input features.

\setlength{\tabcolsep}{6.5pt}
\begin{table*}[t!]
\centering
\begin{tabular}{lccccc|cc}
\toprule
\textbf{Model} & \textbf{Noun} & \textbf{Adjective} & \textbf{Adverb} & \textbf{Verb} & \textbf{Proper Noun} & \textbf{Overall} & \textbf{\iferr{}} \\
\midrule
\mistralsmall      & 0.085 & 0.003 & 0.000 & 0.000 & 0.101 & 0.058 & 0.764 \rlap{(!)}\\
\mistrallarge      & 0.552 & 0.381 & 0.374 & 0.425 & 0.511 & 0.493 & 0.018 \\ \cdashline{1-8}[.4pt/1pt]\noalign{\vskip 0.5ex}
\llamathreesmall   & 0.371 & 0.189 & 0.154 & 0.105 & 0.390 & 0.298 & 0.028 \\
\llamathreelarge   & 0.596 & 0.502 & 0.498 & 0.505 & 0.590 & 0.563 & 0.019 \\
\llamafour         & \textbf{0.610} & \textbf{0.560} & 0.514 & 0.527 & 0.586 & 0.582 & 0.025 \\ \cdashline{1-8}[.4pt/1pt]\noalign{\vskip 0.5ex}
\ayasmall          & 0.403 & 0.334 & 0.340 & 0.377 & 0.335 & 0.379 & 0.050 \\
\ayalarge          & 0.524 & 0.452 & 0.428 & 0.440 & 0.487 & 0.494 & 0.039 \\ \cdashline{1-8}[.4pt/1pt]\noalign{\vskip 0.5ex}
\gemmasmall        & 0.516 & 0.476 & 0.429 & 0.397 & 0.433 & 0.483 & 0.013 \\
\gemmalarge        & 0.609 & 0.534 & \textbf{0.543} & \textbf{0.563} & \textbf{0.594} & \textbf{0.586} & 0.016 \\
\midrule
\textit{average} & 0.474 & 0.381 & 0.364 & 0.371 & 0.447 & - & - \\
\bottomrule
\end{tabular}
\caption{\textbf{Dialect-to-standard translation} -- Accuracy scores for each model across the five most frequent POS tags. The final two columns report the average accuracy across 15 POS (See Table~\ref{tab:full_translation}) and the overall percentage of outputs that failed to follow prompt instructions (\iferr{} rate). 
Bold font shows the \textbf{highest scores}.}
\label{tab:TranslationSelected}
\end{table*}

\section{Results \& Discussion}
\label{sec:results}

We report the task performance for each LLM and part of speech (POS), and across all parts of speeches.
We additionally report the percentage of outputs where LLMs fail to follow prompt instructions (instruction following error rate; \textbf{\iferr{}}). In the Appendix, we show a list of all POS categories and number of instances (Table~\ref{tab:full_statistics}), and results for each POS category (Tables~\ref{tab:full_recognition} and \ref{tab:full_translation}).

\subsection{Judging Translation Candidates}
\label{sec:results-recognition}

Table~\ref{tab:RecognitionSelected} reports macro-F1 scores for the  top five POS categories, together with each model's overall macro-F1 (averaged over all 15 POS tags).

\paragraph{Overall Performance} The results show clear differences across models' performance. In general, for each model family the larger version of the model outperforms the smaller version. \mistrallarge{} obtains the highest overall macro-F1 (0.567), indicating high accuracy across POS tags. Other strong models are \ayalarge{} (0.499) and \gemmasmall{} (0.496), both of which also have very low \iferr{} rates. 

Conversely, \llamafour{}, despite being a newer model, shows lower overall performance (0.368) and the highest \iferr{} (0.438) among all models. While the model can occasionally produce correct outputs, it often fails to follow the requested format and it returns overly long responses rather than the expected single-word answers. In contrast, models such as \mistralsmall{} and \gemmalarge{} also achieve lower overall scores despite relatively low \iferr{} rates, suggesting a gap between instruction-following and actual performance at judging translation candidates. Importantly, several models including \llamathreelarge{}, \ayasmall{}, and \gemmasmall{} show an \iferr{} of zero -- all their outputs followed the prompt requirements, even if the predictions were not always accurate.

\paragraph{POS-Specific Trends} \mistrallarge{} has the best results in all five major POS categories, with especially high results on nouns (0.561) and verbs (0.517). \gemmasmall{} and \ayalarge{} also obtain high macro-F1 scores across most categories, with the former showing strong performance on nouns (0.492), and the latter scoring well on nouns (0.500) and verbs (0.469).

POS difficulty varies across models: nouns (0.438) tend to be the easiest to be correctly classified, while proper nouns (0.373) and adverbs (0.382) often are more challenging. Even models with high overall performance, like \mistrallarge{}, show a drop in correctly judging proper noun pairs.

\paragraph{Qualitative Analysis} We analyzed the classification predictions of the best model (\mistrallarge{}). Overall, we did not find many striking patterns as to which terms get classified correctly or not. Many inflected forms whose inflection status should be obvious due to being identical to the German lemma plus a suffix are misclassified as translations without inflection differences 
(e.g., the Bavarian inflected form \textit{bil\-lign}\footnote{\textit{billig-n} ``cheap''-\textsc{dat\,/\,masc.acc\,/\,def.pl}.} is predicted to be an exact translations of the German lemma \textit{bil\-lig} ``cheap''). Such misclassifications might be influenced by tokenizers that do not usually split such inputs along morpheme lines. Bavarian adjective forms ending with \textit{-schn} (stem ending with \textit{-sch} followed by the inflectional suffix \textit{-n}) almost always get mis-classified (most often as `yes'), but we do not know why these cases should be especially difficult.

Many of the cases where this model fails to follow the instruction of simply outputting a label are words from the Bavarian wiki that are written in non-Latin characters (gold: `no').

\subsection{Dialect-to-Standard Translation}
\label{sec:results-translation}

Table~\ref{tab:TranslationSelected} presents the results for the translation task. Once more, the results indicate high variation in translation quality across model families, sizes, and POS categories.

\paragraph{Overall Performance} Larger and more recent instruction-tuned models consistently outperform smaller ones, showing both substantially higher accuracy scores and lower \iferr{}s -- indicating stronger adherence to the task instructions. It is noteworthy that the best-performing model in the translation task (\gemmalarge{}) is different from the best model in judgment task (\mistrallarge{}). \gemmalarge{} achieves the highest overall accuracy score (0.586), closely followed by \llamafour{} (0.582) and \llamathreelarge{} (0.563). In contrast, \mistralsmall{} performs extremely poorly, with an almost zero overall accuracy. Its extremely high \iferr{} (0.764) further indicates its inability to follow instructions and produce valid outputs.

\paragraph{POS-Specific Trends} Across individual POS categories, \gemmalarge{} shows the best performance for adverbs (0.543), verbs (0.563), and proper nouns (0.594), while \llamafour{} achieves the highest scores for nouns (0.610) and adjectives (0.560). Consistent with results on the judgment task, we find that nouns are the easiest dialect words to translate. However, LLMs show generally higher \iferr{} rates on the translation task. 

\paragraph{Qualitative Analysis} In a few cases, models output a term that the automatic evaluation fails to recognize as correct because of spelling differences sanctioned by German orthography (e.g., \textit{geo\-gra\-fisch} vs.\ \textit{geo\-gra\-phisch}; ``geographic''). When analyzing the outputs of the best-performing translation model (\gemmalarge{}), we observe a range of different error types: Sometimes, the model is just slightly wrong. It might produce a related word with different derivative morphology: \textit{Literaturwissnschoftla} (``literary scholar''; German: \textit{Literaturwissenschaftler}) gets translated as \textit{Literaturwissenschaft} (``literary studies''), or it produces a nonce word that is just one letter away from the real word: \textit{Dreifoitichkeit} (``trinity''; German: \textit{Dreifaltigkeit}) becomes \textit{Dreif\textbf{ä}ltigkeit}. Many words are compound words or they contain prefixes. Frequently, only part of the word is translated correctly: \textit{Iaba\-prifung} (``audit'', lit. ``over+\allowbreak{}test''; German: \textit{Über\-prüfung}) becomes \textit{Jahres\-prüfung} (``yearly test'', lit. ``year+\allowbreak{}test''). However, there are also many translations that are entirely dissimilar: \textit{Vameahrung} (``proliferation''; German: \textit{Vermehrung}) becomes \textit{Wanderung} (``hike''). More generally, there are many cases where a model correctly recognizes a translation pair (`yes') but fails the translation task, or where a model wrongly predicts `no' or `inflected' instead of `yes' and still produces the correct translation. 

\begin{figure}[t!]
    \centering
    \includegraphics[width=1\linewidth]{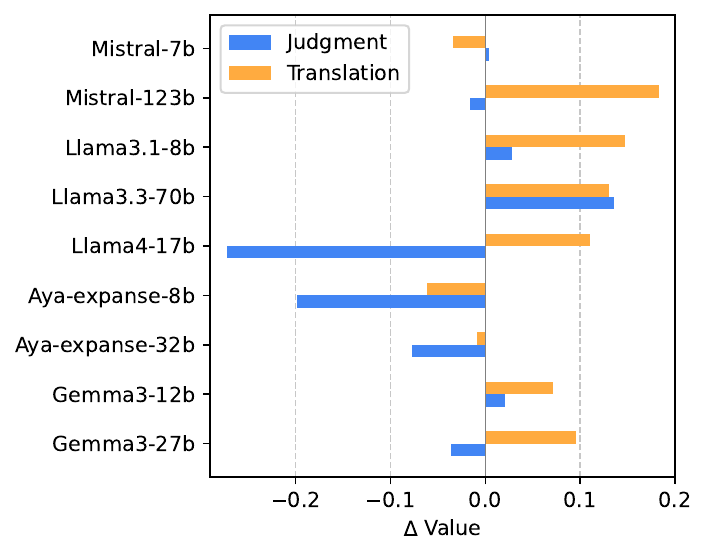}
\caption{\textbf{Context effects} -- Overall changes ($\Delta$) in macro F1 for judging translation candidates (blue bars) and in accuracy for translation (orange bars), measured as the difference between contextualized and non-contextualized prompts.}
    \label{fig:delta_context}
\end{figure}

\subsection{Additional Analyses}
\label{sec:results-additional}

\paragraph{Effect of Context} Figure~\ref{fig:delta_context} shows that adding a short local context (see Section~\ref{sec:dataset-stats}) to the prompts affects model performance to different degrees when judging potential translation candidates (blue bars). 
Among larger models, context generally causes a slight drop in macro-F1 scores. An exception is \llamathreelarge{}, which shows the largest improvement, outperforming \mistrallarge{} and becoming the  best model in this condition (macro F1: 0.564 vs.\ 0.551). In contrast, smaller models show small improvements when adding context but still remain behind their larger counterparts. \llamafour{} shows a substantial drop in its performance, indicating its limited ability to incorporate contextual information effectively. When inspecting the \iferr{} (see left-side Table~\ref{tab:deltas_combined} in the Appendix), we observe that adding context strongly decreases the probability of  following instructions for \mistralsmall{}, \mistrallarge{}, and \ayalarge{}. In contrast, \llamafour{} benefits from contextualization, showing improvements in its ability to follow instructions. 

In the translation task (orange bars), context has a more systematic positive effect: most models show small improvements when additional context is provided. The only exceptions are \mistralsmall{}, \ayasmall{}, and \ayalarge{}, where context slightly reduces performance. Adding context in the translation task slightly worsens instruction-following ability of LLMs (higher \iferr{} rates) in exchange for an overall improved translation performance. This observation is consistent with the pattern we see when comparing smaller to larger models (see discussion below).

\begin{figure}[t!]
    \centering
    \includegraphics[width=1\linewidth]{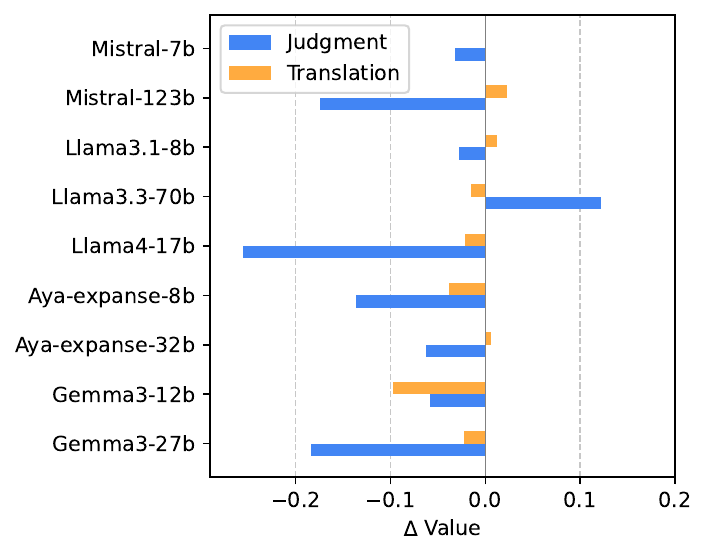}
\caption{\textbf{Language effects} - Overall changes ($\Delta$) in macro F1 for judging translation candidates (blue bars) and in accuracy for translation (orange bars), as the difference between prompts in German and English.}
    \label{fig:delta_language}
\end{figure}

\paragraph{Effect of Prompt Language} For judging the Bavarian--German word pairs, using prompts in German has a negative impact on model performance for all models except \llamathreelarge{}. The change in language also has a strong negative effect on the ability of following instructions, especially for smaller models (see right-side Table~\ref{tab:deltas_combined} in the Appendix). The only strong exception is \llamafour{}, which shows a big improvement in following instructions that written in German. For the translation task, using German prompts only minimally affects model's performance and \iferr{}, with some models showing slight improvements and others slight drops, but without any consistent pattern across model size or family.

\begin{figure}[t!]
    \centering
    \includegraphics[width=0.8\linewidth]{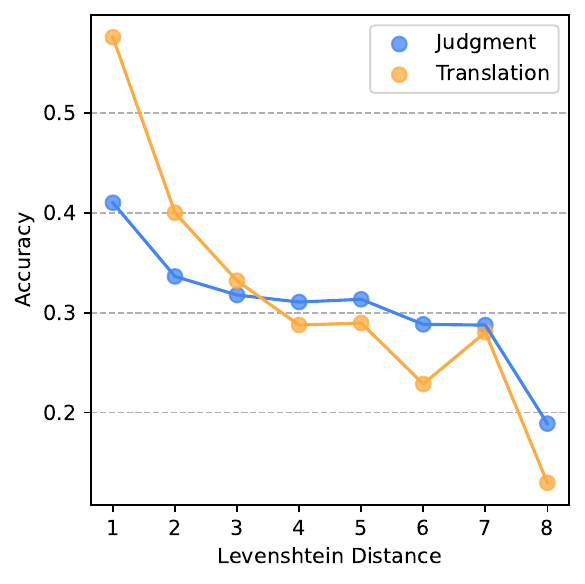}
    \caption{Accuracies for judging translation candidates (excluding the majority class `no') and translations with respect to Levenshtein distances. Results aggregated over models and POS tags.}
    \label{fig:ld-recognition}
\end{figure}

\paragraph{Impact of Levenshtein Distance} Figure~\ref{fig:ld-recognition} shows the accuracy across the first eight Levenshtein distance (LD) values for judging Bavarian--German translation candidates and translating Bavarian words into German, respectively. It is important to mention that for this analysis we excluded the majority class `no' for the judgment task, since models are biased towards predicting `no' (see discussion below). Our results reveal a similar trend in how LD affects LLM performance on both tasks. We find that the model performance deteriorates as word pairs become more lexically distant. Bavarian words like \textit{Basketboispuia} (``basketball player'', German: \textit{Basketballspieler}; $\text{LD}=8$) and \textit{Dochdauntanehmen} (``Subsidiaries'', German: \textit{Tochterunternehmen}; $\text{LD}=6$) 
have a higher LD to their German translation and are more difficult for LLMs. 
Both translations were incorrectly classified  as being unrelated (`no') by \mistrallarge{}.

\begin{table}[t!]
\centering
\small 
\begin{tabular}{|l|ccc|}
\hline
\multicolumn{1}{|c|}{} & \multicolumn{3}{c|}{\textbf{Actual}} \\
\cline{2-4}
\textbf{Predicted} & \textbf{yes} & \textbf{inflected} & \textbf{no} \\
\hline
\textbf{yes} & \multicolumn{1}{c|}{\cellcolor{mistralsmall1}0.76\%} & \multicolumn{1}{c|}{\cellcolor{mistralsmall2}0.41\%} & \multicolumn{1}{c|}{\cellcolor{mistralsmall3}0.52\%} \\
\hline
\textbf{inflected} &  \multicolumn{1}{c|}{\cellcolor{mistralsmall4}5.50\%} & \multicolumn{1}{c|}{\cellcolor{mistralsmall5}7.09\%} & \multicolumn{1}{c|}{\cellcolor{mistralsmall6}3.16\%} \\
\hline
\textbf{no} & \multicolumn{1}{c|}{\cellcolor{mistralsmall7}93.46\%} & \multicolumn{1}{c|}{\cellcolor{mistralsmall8}92.29\%} & \multicolumn{1}{c|}{\cellcolor{mistralsmall9}96.27\%} \\
\hline
\textbf{IF Error} & \multicolumn{1}{c|}{0.28\%} & \multicolumn{1}{c|}{0.21\%} & \multicolumn{1}{c|}{0.05\%}  \\
\hline
\end{tabular}

\caption{Recognition task: Confusion matrix for \mistralsmall{} showing the relative distribution of predictions for each class. F1 scores per class: 0.015 (`yes`), 0.093 (`infl.') and 0.888 (`no').
}
\label{tab:confusion_mistral-small}
\end{table}

\begin{table}[t!]
\centering
\small 
\begin{tabular}{|l|ccc|}
\hline
\multicolumn{1}{|c|}{} & \multicolumn{3}{c|}{\textbf{Actual}} \\
\cline{2-4}
\textbf{Predicted} & \textbf{yes} & \textbf{inflected} & \textbf{no} \\
\hline
\textbf{yes} & \multicolumn{1}{c|}{\cellcolor{mistrallarge1}72.05\%} & \multicolumn{1}{c|}{\cellcolor{mistrallarge2}47.93\%} & \multicolumn{1}{c|}{\cellcolor{mistrallarge3}8.01\%} \\
\hline
\textbf{inflected} &  \multicolumn{1}{c|}{\cellcolor{mistrallarge4}1.02\%} & \multicolumn{1}{c|}{\cellcolor{mistrallarge5}15.97\%} & \multicolumn{1}{c|}{\cellcolor{mistrallarge6}1.62\%} \\
\hline
\textbf{no} & \multicolumn{1}{c|}{\cellcolor{mistrallarge7}25.86\%} & \multicolumn{1}{c|}{\cellcolor{mistrallarge8}34.89\%} & \multicolumn{1}{c|}{\cellcolor{mistrallarge9}87.45\%} \\
\hline
\textbf{IF Error} & \multicolumn{1}{c|}{1.07\%} & \multicolumn{1}{c|}{1.20\%} & \multicolumn{1}{c|}{2.91\%}  \\
\hline
\end{tabular}

\caption{Recognition task: Confusion matrix for \mistrallarge{}. Each column shows the relative distribution of predictions. F1 scores per class: 0.550 (`yes`), 0.234 (`infl.') and 0.902 (`no'). \vspace{-4.4pt}}
\label{tab:confusion_mistral-large}
\end{table}

\paragraph{Impact of Model Size} Our analysis of the ability of LLMs to recognize translation pairs showed that smaller models generally perform worse than their larger variants (Table~\ref{tab:RecognitionSelected}). To gain a deeper understanding of the results, we break down the macro-averaged F1 scores into per-class distributions. Tables~\ref{tab:confusion_mistral-small}~and~\ref{tab:confusion_mistral-large} show the confusion matrices (\%) for \mistralsmall{} and \mistrallarge{}. We can see that \mistralsmall{} has a strong bias for predicting `no'. When moving to the larger \mistrallarge{} model, however, we find that the predictions shift from `no' (before: 93.46\% and 92.29\% of `yes' and `inflected' instances) towards the other two classes (after: 25.86\% and 34.8\% of `yes' and `inflected' instances). We also observe a slightly higher \iferr{} rate for larger models. Please refer to Appendix~\ref{sec:appendix} for further analyses.

\section{Conclusion}
\label{sec:conclusion}

We introduce \dialemma{}, a novel dialect annotation framework and build a dataset consisting of German lemmas and Bavarian variants. We conduct a systematic analysis of the ability of LLMs on two tasks: judging whether superficially similar dialect--standard word pairs are translations of each other, and translating dialect variants into their standard-language counterparts. Our results show that their performance varies by model size and the part-of-speech of the input word. In future work, we plan to 1) fine-tune LLMs for identifying translation pairs and as translators and 2) extend our evaluation protocol to extrinsic evaluation on downstream tasks. We will release our code and dataset for future uptake (\href{https://creativecommons.org/licenses/by-sa/4.0/deed.en}{CC BY-SA 4.0}).

\section*{Limitations}

The main limitation of our study is its focus on Bavarian as a single case study. While this allowed for a more precise and controlled investigation, we see a lot of potential to extend this approach to other low-resource language scenarios.

Even though we added contextual information (e.g., short Wikipedia extracts) to better align the model setup with human annotation conditions, we acknowledge that context can influence model performance in more complex ways than those explored here. Future work should systematically analyze the role of contextual information in identifying dialect--standard translation pairs and generating dialect-to-standard translations. Finally, the focus of this work lies on spelling variation of cognates with the standard language. By definition of the Levenshtein distance, this excludes other differences, such as regional word choices. 

We used SpaCy for lemmatization and POS tagging. Although the tool works well (the model documentation reports an accuracy of 98\,\% for both tasks),\footnote{\url{https://spacy.io/models/de\#de_core_news_lg-accuracy}} it is not perfect, occasionally producing an incorrect lemma or mis-tagging words. Inspecting our data, we found that this might especially be the case for adjectives that wrongly were tagged as adverbs.

\section*{Ethical considerations}

We see no ethical issues related to this work. All experiments were conducted with publicly available data and open-source software, and we have made all of our code and linguistic resources openly available for reproducibility. All annotators were fairly compensated for their work. One of the three annotators was hired as a research assistant and paid according to standard national wages. The other two annotators contributed their annotations as part of their thesis work. We transparently communicated the use of their annotations and obtained explicit consent prior to the start of the project.

\section*{Use of AI Assistants} The authors acknowledge the use of ChatGPT for correcting grammatical errors, enhancing the coherence of the final manuscripts, and providing assistance with prompt engineering.

\section*{Acknowledgments}
We thank Lisa Miller and Miriam Winkler for their help with annotating the dataset. This research is in parts supported by European Research Council (ERC) Consolidator Grant DIALECT 101043235.

\bibliography{custom,anthology}

\appendix

\begin{table}[h!]
    \centering
    \resizebox{\linewidth}{!}{
    \begin{tabular}{l rrr r} 
    \toprule
& \multicolumn{3}{c}{\textbf{Judgment}} & \\ \cmidrule(lr){2-4}
\textbf{Part-of-speech} & \textbf{Yes} & \textbf{Inflected} & \textbf{No} & \textbf{Translation} \\
\midrule 
Noun & 6,720 & 2,670 & 28,480 & 6,564 \\
Adjective & 1,358 & 3,066 & 5,496 & 1,325 \\
Adverb & 1,157 & -- & 2,783 & 1,126 \\
Verb & 934 & 1,182 & 6,214 & 916 \\
Proper Noun & 574 & 86 & 34,430 & 556 \\ \cdashline{1-5}[.4pt/1pt]\noalign{\vskip 0.5ex}
Adposition & 148 & -- & 342 & 144 \\
Numeral & 45 & -- & 175 & 45 \\
Sub. Conjunction & 35 & -- & 185 & 34 \\
Determiner & 34 & 37 & 139 & 34 \\
Auxiliary & 8 & 17 & 75 & 8 \\
Pronoun & 8 & 9 & 153 & 8 \\
Coord. Conjunction & 15 & -- & 65 & 7 \\
Other (X) & 5 & 3 & 3,012 & 5 \\
Interjection & 3 & -- & 7 & 3 \\ 
Particle & -- & -- & 30 & -- \\
\cdashline{1-5}[.4pt/1pt]\noalign{\vskip 0.5ex}

Total & 11,044 & 7,070 & 81,586 & 10,775 \\ 

\bottomrule
    \end{tabular}
    }
    \caption{Number of test instances in each part-of-speech class for both tasks.}
    \label{tab:full_statistics}
\end{table}

\section{Full Dataset Statistics}

Table~\ref{tab:full_statistics} shows number of instances for the datasets used in both word-level dialect tasks, which are introduced in Section~\ref{sec:dataset-tasks}. In addition to total number of instances, we also show number of instances for each part of speech.

\section{Full Results}
\label{sec:appendix}

Table~\ref{tab:full_recognition} contains the detailed results for the task of judging whether a Bavarian word is a translation of a German lemma. Table~\ref{tab:full_translation} shows the detailed results for the translation task. 

\begin{table}[t!]
\small
    \centering
\begin{tabular}{|l|ccc|}
\hline
\multicolumn{1}{|c|}{} & \multicolumn{3}{c|}{\textbf{Actual}} \\
\cline{2-4}
\textbf{Predicted} & \textbf{yes} & \textbf{inflected} & \textbf{no} \\
\hline
\textbf{yes} & \multicolumn{1}{c|}{\cellcolor{llama3.11}6.82\%} & \multicolumn{1}{c|}{\cellcolor{llama3.12}3.78\%} & \multicolumn{1}{c|}{\cellcolor{llama3.13}0.80\%} \\
\hline
\textbf{inflected} &  \multicolumn{1}{c|}{\cellcolor{llama3.14}14.17\%} & \multicolumn{1}{c|}{\cellcolor{llama3.15}15.28\%} & \multicolumn{1}{c|}{\cellcolor{llama3.16}2.87\%} \\
\hline
\textbf{no} & \multicolumn{1}{c|}{\cellcolor{llama3.17}78.95\%} & \multicolumn{1}{c|}{\cellcolor{llama3.18}80.89\%} & \multicolumn{1}{c|}{\cellcolor{llama3.19}96.00\%} \\
\hline
\textbf{IF Error} & \multicolumn{1}{c|}{0.06\%} & \multicolumn{1}{c|}{0.06\%} & \multicolumn{1}{c|}{0.33\%}  \\
\hline
\end{tabular}
    \caption{Confusion matrix for \llamathreesmall{}. F1 scores per class: 0.118 (`yes`), 0.179 (`infl.'), and 0.898 (`no').}
    \label{tab:conf_llamasmall}
\end{table}

\begin{table}[t!]
    \centering
    \small
\begin{tabular}{|l|ccc|}
\hline
\multicolumn{1}{|c|}{} & \multicolumn{3}{c|}{\textbf{Actual}} \\
\cline{2-4}
\textbf{Predicted} & \textbf{yes} & \textbf{inflected} & \textbf{no} \\
\hline
\textbf{yes} & \multicolumn{1}{c|}{\cellcolor{llama3.31}86.07\%} & \multicolumn{1}{c|}{\cellcolor{llama3.32}77.57\%} & \multicolumn{1}{c|}{\cellcolor{llama3.33}24.75\%} \\
\hline
\textbf{inflected} &  \multicolumn{1}{c|}{\cellcolor{llama3.34}0.16\%} & \multicolumn{1}{c|}{\cellcolor{llama3.35}1.94\%} & \multicolumn{1}{c|}{\cellcolor{llama3.36}0.51\%} \\
\hline
\textbf{no} & \multicolumn{1}{c|}{\cellcolor{llama3.37}13.76\%} & \multicolumn{1}{c|}{\cellcolor{llama3.38}20.50\%} & \multicolumn{1}{c|}{\cellcolor{llama3.39}74.73\%} \\
\hline
\textbf{IF Error} & \multicolumn{1}{c|}{0.00\%} & \multicolumn{1}{c|}{0.00\%} & \multicolumn{1}{c|}{0.01\%}  \\
\hline
\end{tabular}
    \caption{Conf. matrix for \llamathreelarge{}. F1 scores per class: 0.411 (`yes`), 0.036 (`infl.'), and 0.838 (`no').}
    \label{tab:conf_llamalarge}
\end{table}

\begin{table}[t!]
    \centering
    \small
\begin{tabular}{|l|ccc|}
\hline
\multicolumn{1}{|c|}{} & \multicolumn{3}{c|}{\textbf{Actual}} \\
\cline{2-4}
\textbf{Predicted} & \textbf{yes} & \textbf{inflected} & \textbf{no} \\
\hline
\textbf{yes} & \multicolumn{1}{c|}{\cellcolor{llama41}5.94\%} & \multicolumn{1}{c|}{\cellcolor{llama42}0.75\%} & \multicolumn{1}{c|}{\cellcolor{llama43}0.42\%} \\
\hline
\textbf{inflected} &  \multicolumn{1}{c|}{\cellcolor{llama44}76.37\%} & \multicolumn{1}{c|}{\cellcolor{llama45}82.89\%} & \multicolumn{1}{c|}{\cellcolor{llama46}35.60\%} \\
\hline
\textbf{no} & \multicolumn{1}{c|}{\cellcolor{llama47}2.36\%} & \multicolumn{1}{c|}{\cellcolor{llama48}1.36\%} & \multicolumn{1}{c|}{\cellcolor{llama49}13.78\%} \\
\hline
\textbf{IF Error} & \multicolumn{1}{c|}{\cellcolor{llama410} 15.33\%} & \multicolumn{1}{c|}{\cellcolor{llama411}15.01\%} & \multicolumn{1}{c|}{\cellcolor{llama412} 50.20\%}  \\
\hline
\end{tabular}
    \caption{Confusion matrix for \llamafour{}. F1 scores per class: 0.108 (`yes`), 0.233 (`infl.'), and 0.763 (`no').}
    \label{tab:conf_llamafour}
\end{table}

\paragraph{Judgment task} Confusion matrices for \texttt{Llama}, \texttt{Gemma} and \texttt{Aya} models are shown in Tables~\ref{tab:conf_llamasmall}-\ref{tab:conf_gemmalarge}. We find that \texttt{Llama} and \texttt{Mistral} models show the same pattern: smaller LLMs are biased towards predicting `no', while their larger versions can identify dialect variants (`yes', `inflected') more accurately. \llamafour{} stands out with a much higher \textbf{IF~Error} rate. \texttt{Gemma} and \texttt{Aya} models already show relatively strong results in their smaller variants. Detailed results are shown in Table~\ref{tab:precision_recall}.

\begin{table}[t!]
    \centering
    \small
\begin{tabular}{|l|ccc|}
\hline
\multicolumn{1}{|c|}{} & \multicolumn{3}{c|}{\textbf{Actual}} \\
\cline{2-4}
\textbf{Predicted} & \textbf{yes} & \textbf{inflected} & \textbf{no} \\
\hline
\textbf{yes} & \multicolumn{1}{c|}{\cellcolor{ayasmall1}38.68\%} & \multicolumn{1}{c|}{\cellcolor{ayasmall2}37.75\%} & \multicolumn{1}{c|}{\cellcolor{ayasmall3}4.91\%} \\
\hline
\textbf{inflected} &  \multicolumn{1}{c|}{\cellcolor{ayasmall4}0.32\%} & \multicolumn{1}{c|}{\cellcolor{ayasmall5}0.38\%} & \multicolumn{1}{c|}{\cellcolor{ayasmall6}0.03\%} \\
\hline
\textbf{no} & \multicolumn{1}{c|}{\cellcolor{ayasmall7}61.00\%} & \multicolumn{1}{c|}{\cellcolor{ayasmall8}61.87\%} & \multicolumn{1}{c|}{\cellcolor{ayasmall9}95.05\%} \\
\hline
\textbf{IF Error} & \multicolumn{1}{c|}{0.00\%} & \multicolumn{1}{c|}{0.00\%} & \multicolumn{1}{c|}{0.00\%}  \\
\hline
\end{tabular}
    \caption{Conf. matrix for \ayasmall{}. F1 scores per class: 0.388 (`yes`), 0.008 (`infl.'), and 0.911 (`no').}
    \label{tab:conf_ayasmall}
\end{table}

\begin{table}[t!]
    \centering
    \small
\begin{tabular}{|l|ccc|}
\hline
\multicolumn{1}{|c|}{} & \multicolumn{3}{c|}{\textbf{Actual}} \\
\cline{2-4}
\textbf{Predicted} & \textbf{yes} & \textbf{inflected} & \textbf{no} \\
\hline
\textbf{yes} & \multicolumn{1}{c|}{\cellcolor{ayalarge1}46.73\%} & \multicolumn{1}{c|}{\cellcolor{ayalarge2}34.10\%} & \multicolumn{1}{c|}{\cellcolor{ayalarge3}7.07\%} \\
\hline
\textbf{inflected} &  \multicolumn{1}{c|}{\cellcolor{ayalarge4}7.43\%} & \multicolumn{1}{c|}{\cellcolor{ayalarge5}12.91\%} & \multicolumn{1}{c|}{\cellcolor{ayalarge6}2.14\%} \\
\hline
\textbf{no} & \multicolumn{1}{c|}{\cellcolor{ayalarge7}44.17\%} & \multicolumn{1}{c|}{\cellcolor{ayalarge8}51.60\%} & \multicolumn{1}{c|}{\cellcolor{ayalarge9}90.47\%} \\
\hline
\textbf{IF Error} & \multicolumn{1}{c|}{\cellcolor{ayalarge10}1.67\%} & \multicolumn{1}{c|}{\cellcolor{ayalarge11}1.39\%} & \multicolumn{1}{c|}{\cellcolor{ayalarge12}0.32\%}  \\
\hline
\end{tabular}
    \caption{Conf. matr. for \ayalarge{}. F1 scores per class: 0.423 (`yes`), 0.173 (`infl.'), and 0.901 (`no').}
    \label{tab:conf_ayalarge}
\end{table}

\begin{table}[t!]
    \centering
    \small
\begin{tabular}{|l|ccc|}
\hline
\multicolumn{1}{|c|}{} & \multicolumn{3}{c|}{\textbf{Actual}} \\
\cline{2-4}
\textbf{Predicted} & \textbf{yes} & \textbf{inflected} & \textbf{no} \\
\hline
\textbf{yes} & \multicolumn{1}{c|}{\cellcolor{gemmasmall1}60.28\%} & \multicolumn{1}{c|}{\cellcolor{gemmasmall2}48.60\%} & \multicolumn{1}{c|}{\cellcolor{gemmasmall3}13.86\%} \\
\hline
\textbf{inflected} &  \multicolumn{1}{c|}{\cellcolor{gemmasmall4}6.90\%} & \multicolumn{1}{c|}{\cellcolor{gemmasmall5}18.46\%} & \multicolumn{1}{c|}{\cellcolor{gemmasmall6}3.94\%} \\
\hline
\textbf{no} & \multicolumn{1}{c|}{\cellcolor{gemmasmall7}32.82\%} & \multicolumn{1}{c|}{\cellcolor{gemmasmall8}32.94\%} & \multicolumn{1}{c|}{\cellcolor{gemmasmall9}82.20\%} \\
\hline
\textbf{IF Error} & \multicolumn{1}{c|}{0.00\%} & \multicolumn{1}{c|}{0.00\%} & \multicolumn{1}{c|}{0.00\%}  \\
\hline
\end{tabular}
    \caption{Confusion matrix for \gemmasmall{}. F1 scores per class: 0.410 (`yes`), 0.211 (`infl.'), and 0.868 (`no').}
    \label{tab:conf_gemmasmall}
\end{table}

\begin{table}[t!]
    \centering
    \small
\begin{tabular}{|l|ccc|}
\hline
\multicolumn{1}{|c|}{} & \multicolumn{3}{c|}{\textbf{Actual}} \\
\cline{2-4}
\textbf{Predicted} & \textbf{yes} & \textbf{inflected} & \textbf{no} \\
\hline
\textbf{yes} & \multicolumn{1}{c|}{\cellcolor{gemmalarge1}78.24\%} & \multicolumn{1}{c|}{\cellcolor{gemmalarge2}49.02\%} & \multicolumn{1}{c|}{\cellcolor{gemmalarge3}43.60\%} \\
\hline
\textbf{inflected} &  \multicolumn{1}{c|}{\cellcolor{gemmalarge4}11.05\%} & \multicolumn{1}{c|}{\cellcolor{gemmalarge5}42.97\%} & \multicolumn{1}{c|}{\cellcolor{gemmalarge6}8.99\%} \\
\hline
\textbf{no} & \multicolumn{1}{c|}{\cellcolor{gemmalarge7}10.71\%} & \multicolumn{1}{c|}{\cellcolor{gemmalarge8}8.01\%} & \multicolumn{1}{c|}{\cellcolor{gemmalarge9}47.32\%} \\
\hline
\textbf{IF Error} & \multicolumn{1}{c|}{0.00\%} & \multicolumn{1}{c|}{0.00\%} & \multicolumn{1}{c|}{0.09\%}  \\
\hline
\end{tabular}
    \caption{Confusion matrix for \gemmalarge{}. F1 scores per class: 0.294 (`yes`), 0.326 (`infl.'), and 0.633 (`no').}
    \label{tab:conf_gemmalarge}
\end{table}

\section{Prompt selection}
\label{appendix:prompts}

We include the prompts we included in the prompt pool (Section~\ref{sec:method-prompts}) in Tables \ref{tab:recognition_prompt_templates}  and \ref{tab:translation_prompt_templates}.
For the test set, we only use the best prompt per model (highlighted rows).
This prompt selection matters, as the LLMs we test are all sensitive to the way prompts are phrased:
Table~\ref{tbl:promptVar_recognition} illustrates this for the word pair judgments, and Table~\ref{tbl:promptVar_translation} for the word-level dialect translation task.

We additionally evaluate versions of the best prompts that include a usage example of the word in context (Tables \ref{tab:recognition_w_context_prompts} and \ref{tab:translation_w_context_prompts}) and German-language versions (Tables \ref{tab:recognition_german_prompts} and \ref{tab:translation_german_prompts}).
The results for these modified prompts are in Table~\ref{tab:deltas_combined}.

\setlength{\tabcolsep}{7.5pt}
\begin{table*}[t!]
    \centering
    \small 
    \begin{tabular}{lcccc ccccc}
    \toprule
    & \multicolumn{3}{c}{\textbf{`yes'}} & \multicolumn{3}{c}{\textbf{`inflected'}} & \multicolumn{3}{c}{\textbf{`no'}} \\  \cmidrule(lr){2-4}\cmidrule(lr){5-7}\cmidrule(lr){8-10} 
Model & Precision & Recall & F1 & Precision & Recall & F1 & Precision & Recall & F1 \\ 
\midrule
\mistralsmall{} & 0.157 & 0.008 & 0.015 & 0.136 & 0.071 & 0.093 & 0.823 & 0.963 & 0.888 \\
\mistrallarge{} & 0.445 & 0.720 & 0.550 & 0.440 & 0.160 & 0.234 & 0.931 & 0.875 & 0.902 \\
\llamathreesmall{} & 0.450 & 0.068 & 0.118 & 0.217 & 0.153 & 0.179 & 0.844 & 0.96 & 0.898 \\
\llamathreelarge{} & 0.270 & 0.861 & 0.411 & 0.240 & 0.019 & 0.036 & 0.954 & 0.747 & 0.838 \\
\llamafour{} & 0.623 & 0.059 & 0.108 & 0.135 & 0.829 & 0.233 & 0.944 & 0.640 & 0.763 \\
\ayasmall{} & 0.390 & 0.387 & 0.388 & 0.300 & 0.004 & 0.008 & 0.875 & 0.951 & 0.911 \\
\ayalarge{} & 0.387 & 0.467 & 0.423 & 0.263 & 0.129 & 0.173 & 0.894 & 0.908 & 0.901 \\
\gemmasmall{} & 0.311 & 0.603 & 0.410 & 0.247 & 0.185 & 0.211 & 0.918 & 0.822 & 0.868 \\
\gemmalarge{} & 0.181 & 0.782 & 0.294 & 0.262 & 0.430 & 0.326 & 0.957 & 0.473 & 0.633 \\
\bottomrule
    \end{tabular}
    \caption{Precision, Recall and F1 scores of individual classes in our dialect judgment task. Instances where LLMs fail to follow instructions are considered the same as predicting `no'.}
    \label{tab:precision_recall}
\end{table*}

\vspace{4\baselineskip}

\setlength{\tabcolsep}{3.5pt}
\begin{table*}
\centering
\small
\begin{tabular}{lcccc| cc | ccc |cc | cc}
\toprule
& \multicolumn{4}{c|}{\textbf{\texttt{Baselines}}} & \multicolumn{2}{c|}{\textbf{\texttt{Mistral}}} & \multicolumn{3}{c|}{\textbf{\texttt{Llama}}} & \multicolumn{2}{c|}{\textbf{\texttt{Aya-expanse}}} & \multicolumn{2}{c}{\textbf{\texttt{Gemma}}}\\

& \textbf{Rand.} & \textbf{\texttt{LD}} & \textbf{\texttt{Maj.}} & \textbf{\texttt{Log.Reg.}} & \textbf{\texttt{7b}} & \textbf{\texttt{123b}} & \textbf{\texttt{3.1-8b}} & \textbf{\texttt{3.3-70b}} & \textbf{\texttt{4-17b}} & \textbf{\texttt{8b}} & \textbf{\texttt{32b}} & \textbf{\texttt{12b}} & \textbf{\texttt{27b}} \\

\midrule
\textbf{Noun}   & 0.268 & 0.355 & 0.286 & 0.351 & 0.292 & \textbf{0.561} & 0.398 & 0.444 & 0.354 & 0.447 & 0.500 & 0.492 & 0.453 \\
\textbf{Adjective}  & 0.309 & 0.307 & 0.238 & 0.310  & 0.278 & \textbf{0.488} & 0.305 & 0.369 & 0.394 & 0.332 & 0.411 & 0.444 & 0.431  \\
\textbf{Adverb}  & 0.257 & 0.331 & 0.276 & 0.270 & 0.270 & \textbf{0.484} & 0.300 & 0.379 & 0.237 & 0.392 & 0.442 & 0.441 & 0.288  \\
\textbf{Verb}   & 0.280 & 0.307 & 0.285 & 0.344 & 0.335 & \textbf{0.517} & 0.351 & 0.340 & 0.338 & 0.390 & 0.469 & 0.430 & 0.323 \\
\textbf{Proper Noun}  & 0.174 & 0.202 & 0.330 & 0.312 & 0.330 & \textbf{0.473} & 0.383 & 0.359 & 0.309 & 0.436 & 0.414 & 0.390 & 0.261  \\ \cdashline{1-14}[.4pt/1pt]\noalign{\vskip 0.5ex}
\textbf{Adposition}    & 0.262 & 0.264 & 0.274 & 0.257 & 0.277 & \textbf{0.519} & 0.284 & 0.422 & 0.261 & 0.421 & 0.469 & 0.489 & 0.285 \\
\textbf{Numeral}    & 0.242 & 0.336 & 0.295 & 0.292 & 0.313 & \textbf{0.487} & 0.352 & 0.365 & 0.260 & 0.471 & 0.484 & 0.454 & 0.326 \\
\textbf{Sub. Conj.}  & 0.225 & 0.142 & 0.305 & 0.305 & 0.310 & \textbf{0.453} & 0.301 & 0.249 & 0.234 & 0.417 & 0.445 & 0.409 & 0.159 \\
\textbf{Determiner}    & 0.235 & 0.136 & 0.266 & 0.260 & 0.343 & \textbf{0.479} & 0.382 & 0.225 & 0.264 & 0.377 & 0.406 & 0.421 & 0.226  \\
\textbf{Auxiliary}    & 0.356 & 0.076 & 0.286 & 0.322 & 0.350 & \textbf{0.552} & 0.473 & 0.175 & 0.398 & 0.463 & 0.506 & 0.475 & 0.226 \\ 
\textbf{Pronoun}   & 0.187 & 0.091 & 0.316 & 0.296 & 0.310 & 0.409 & 0.412 & 0.219 & 0.263 & 0.365 & 0.387 & \textbf{0.415} & 0.170 \\
\textbf{Coord. Conj.}  & 0.184 & 0.178 & 0.299 & 0.291 & 0.247 & \textbf{0.437} & 0.301 & 0.225 & 0.211 & 0.421 & 0.407 & 0.351 & 0.186  \\
\textbf{Other (X)}      & 0.169 & 0.093 & 0.333 & 0.314 & 0.323 & \textbf{0.356} & 0.332 & 0.294 & 0.308 & 0.331 & 0.336 & 0.333 & 0.197 \\
\textbf{Interjection}   & 0.300 & 0.154 & 0.275 & 0.083 & 0.061 & 0.250 & 0.275 & 0.061 & 0.083 & \textbf{0.593} & 0.330 & 0.275 & 0.154 \\
\bottomrule
\end{tabular}
\caption{\textbf{Judging potential German--Bavarian translation candidates} -- Macro F1 scores for each model across all 15 POS tags. In bold font we highlight the \textbf{highest score} for each POS.}
\label{tab:full_recognition}

\vspace{4\baselineskip}

\setlength{\tabcolsep}{8pt}
\small
\begin{tabular}{lcc|ccc|cc|cc}
\toprule
 & \multicolumn{2}{c|}{\textbf{\texttt{Mistral}}} & \multicolumn{3}{c|}{\textbf{\texttt{Llama}}} & \multicolumn{2}{c|}{\textbf{\texttt{Aya-expanse}}} & \multicolumn{2}{c}{\textbf{\texttt{Gemma}}} \\
& \textbf{\texttt{7b}} & \textbf{\texttt{123b}} & \textbf{\texttt{3.1-8b}} & \textbf{\texttt{3.3-70b}} & \textbf{\texttt{4-17b}} & \textbf{\texttt{8b}} & \textbf{\texttt{32b}} & \textbf{\texttt{12b}} & \textbf{\texttt{27b}} \\
\midrule
\textbf{Noun}               & 0.085 & 0.552 & 0.371 & 0.596 & \textbf{0.610} & 0.403 & 0.524 & 0.516 & 0.609 \\
\textbf{Adjective}          & 0.003 & 0.381 & 0.189 & 0.502 & \textbf{0.560} & 0.334 & 0.452 & 0.476 & 0.534 \\
\textbf{Adverb}             & 0.000 & 0.374 & 0.154 & 0.498 & 0.514 & 0.340 & 0.428 & 0.429 & \textbf{0.543} \\
\textbf{Verb}               & 0.000 & 0.425 & 0.105 & 0.505 & 0.527 & 0.377 & 0.440 & 0.397 & \textbf{0.563} \\
\textbf{Proper Noun}        & 0.101 & 0.511 & 0.390 & 0.590 & 0.586 & 0.335 & 0.487 & 0.433 & \textbf{0.594} \\ \cdashline{1-10}[.4pt/1pt]\noalign{\vskip 0.5ex}
\textbf{Adposition}         & 0.000 & 0.403 & 0.160 & 0.479 & 0.472 & 0.326 & 0.479 & 0.354 & \textbf{0.556} \\
\textbf{Numeral}            & 0.000 & 0.333 & 0.156 & \textbf{0.689} & 0.667 & 0.378 & \textbf{0.689} & 0.489 & \textbf{0.689} \\
\textbf{Sub. Conjunction}   & 0.000 & 0.118 & 0.000 & 0.118 & \textbf{0.294} & 0.147 & 0.206 & 0.324 & 0.294 \\
\textbf{Determiner}         & 0.000 & 0.206 & 0.088 & 0.265 & \textbf{0.412} & 0.294 & 0.324 & 0.235 & 0.471 \\
\textbf{Auxiliary}          & 0.000 & 0.750 & 0.375 & 0.875 & \textbf{1.000} & 0.625 & 0.625 & 0.750 & 0.875 \\
\textbf{Pronoun}            & 0.000 & 0.125 & 0.000 & \textbf{0.250} & 0.125 & 0.125 & \textbf{0.250} & 0.000 & \textbf{0.250} \\
\textbf{Coord. Conjunction} & 0.000 & 0.429 & 0.286 & \textbf{0.857} & 0.714 & 0.429 & 0.571 & 0.714 & 0.714 \\
\textbf{Other (X)}          & 0.200 & 0.200 & 0.200 & \textbf{0.400} & \textbf{0.400} & 0.200 & 0.200 & 0.200 & 0.200 \\
\textbf{Interjection}       & 0.000 & 0.000 & 0.000 & 0.333 & 0.000 & 0.000 & 0.000 & 0.333 & \textbf{0.667} \\
\bottomrule
\end{tabular}
    \caption{\textbf{Dialect-to-standard translation} -- Accuracy scores for each model across all 15 POS tags. In bold font we highlight the \textbf{highest score} for each POS.} 
\label{tab:full_translation}\end{table*}

\renewcommand{\arraystretch}{1.5}
\setlength{\tabcolsep}{3pt}

\begin{table*}[h!]
    \centering
    \footnotesize 

\begin{tabular}{l p{0.95\textwidth}}
\toprule
ID & Prompt template \\ \midrule 

0 & 
\begin{minipage}[t]{0.95\textwidth}
\begin{minted}[breaklines,breaksymbolleft=]{text}
Classify the Bavarian term 'term_bar' in relation to the Standard German term 'term_de'. Return 'yes' if it is an exact dialectal translation, 'inflected' if it is an inflected dialectal translation of it, or 'no' if neither applies. Do not say any other word.
\end{minted}
\end{minipage} \\ \hline 

1 & 
\begin{minipage}[t]{0.95\textwidth}
\begin{minted}[breaklines,breaksymbolleft=]{text}
Evaluate the relationship between 'term_bar' (Bavarian) and 'term_de' (Standard German). Respond with 'yes' for a dialectal match, 'inflected' for a dialectal inflectional variant, and 'no' if neither. Do not say any other word.
\end{minted}
\end{minipage} \\ \hline

\rowcolor{blanchedalmond} 2 & 
\begin{minipage}[t]{0.95\textwidth}
\begin{minted}[breaklines,breaksymbolleft=]{text}
Is the Bavarian term 'term_bar' an exact dialectal variant ('yes'), a dialectal morphological inflection ('inflected'), or not a dialectal variant ('no') of 'term_de' in Standard German? Return only "yes", "inflected", or "no".
\end{minted}
\end{minipage} \\ \hline

\rowcolor{blanchedalmond} 3 & 
\begin{minipage}[t]{0.95\textwidth}
\begin{minted}[breaklines,breaksymbolleft=]{text}
Compare 'term_bar' (Bavarian) to 'term_de' (Standard German). Answer with 'yes' if it's a direct dialectal translation, 'inflected' if it's a dialectal inflected form, or 'no' if neither applies. Do not say any other word.
\end{minted}
\end{minipage} \\ \hline

4 & 
\begin{minipage}[t]{0.95\textwidth}
\begin{minted}[breaklines,breaksymbolleft=]{text}
Categorize the Bavarian term 'term_bar' with respect to the German term 'term_de':
- 'yes' = dialectal variant
- 'inflected' = morphologically related
- 'no' = otherwise
Return only one label.
\end{minted}
\end{minipage} \\ \hline

\rowcolor{blanchedalmond} 5 & 
\begin{minipage}[t]{0.95\textwidth}
\begin{minted}[breaklines,breaksymbolleft=]{text}
Task: Is the Bavarian term: "term_bar" a correct dialectal variant of the German term: "term_de"? Follow the given annotation guidelines.
Guidelines:
- yes: The candidate is an exact dialectal variation of the Standard German word.
- inflected: The candidate is a morphologically inflected variant of the German word.
- no: None of the two applies.
Return only "yes", "inflected", or "no".
\end{minted}
\end{minipage} \\ \hline

\rowcolor{blanchedalmond} 6 & 
\begin{minipage}[t]{0.95\textwidth}
\begin{minted}[breaklines,breaksymbolleft=]{text}
Classify the Bavarian term 'term_bar' with respect to the Standard German term 'term_de'. Return exactly one of the following:
- 'yes' if it is an exact dialectal variant
- 'inflected' if it is a morphologically inflected variant
- 'no' otherwise
\end{minted}
\end{minipage} \\ \hline

7 & 
\begin{minipage}[t]{0.95\textwidth}
\begin{minted}[breaklines,breaksymbolleft=]{text}
Task: Label the relationship between term_bar (Bavarian) and term_de (German) using these rules:
- 'yes': direct dialectal equivalent
- 'inflected': same lemma, different form
- 'no': if neither applies
\end{minted}
\end{minipage} \\ \hline

\rowcolor{blanchedalmond} 8 & 
\begin{minipage}[t]{0.95\textwidth}
\begin{minted}[breaklines,breaksymbolleft=]{text}
Is the Bavarian term 'term_bar':
- a dialectal translation of the German term 'term_de' → 'yes'
- an inflection of the German term 'term_de' → 'inflected'
- not a variant or inflected variant of the German term 'term_de' → 'no'
Answer only with 'yes', 'inflected', or 'no'.
\end{minted}
\end{minipage} \\ \hline

9 & 
\begin{minipage}[t]{0.95\textwidth}
\begin{minted}[breaklines,breaksymbolleft=]{text}
Compare the Bavarian word 'term_bar' against the Standard German 'term_de'.
Select only one label:
- 'yes' for dialectal variant
- 'inflected' for inflectional variant
- 'no' otherwise
\end{minted}
\end{minipage} \\ \hline

10 & 
\begin{minipage}[t]{0.95\textwidth}
\begin{minted}[breaklines,breaksymbolleft=]{text}
Task: Is the Bavarian term: "term_bar" a correct dialectal variant of the German term: "term_de"? Follow the given annotation guidelines.
Guidelines:
- yes: The candidate is an exact dialectal variant of the Standard German word.
- inflected: The candidate is an inflected dialectal variant of the German word.
- no: None of the two applies.
Return only "yes", "inflected", or "no".
\end{minted}
\end{minipage} \\ \hline

11 & 
\begin{minipage}[t]{0.95\textwidth}
\begin{minted}[breaklines,breaksymbolleft=]{text}
Is the Bavarian term 'term_bar':
- a dialectal translation of the German term 'term_de' → 'yes'
- an inflected dialectal translation of the German term 'term_de' → 'inflected'
- not an (inflected or uninflected) dialectal translation of the German term 'term_de' → 'no'
Answer only with 'yes', 'inflected', or 'no'.
\end{minted}
\end{minipage} \\ \hline

12 & 
\begin{minipage}[t]{0.95\textwidth}
\begin{minted}[breaklines,breaksymbolleft=]{text}
Classify the Bavarian term 'term_bar' with respect to the Standard German term 'term_de'. Return exactly one of the following:
- 'yes' if it is an exact dialectal variant
- 'inflected' if it is a morphologically inflected dialectal variant
- 'no' otherwise
\end{minted}
\end{minipage} \\  \bottomrule

\end{tabular}

    \caption{\textbf{Translation candidate judgment} -- \colorbox{blanchedalmond}{Best prompts}: \mistralsmall{} (2), \mistrallarge{} (8), \llamathreesmall{} (5), \llamathreelarge{} (5), \llamafour{} (8), \ayasmall{} (3), \ayalarge{} (5), \gemmasmall{} (8), \gemmalarge{} (6).}
    \label{tab:recognition_prompt_templates}
\end{table*}

\renewcommand{\arraystretch}{1.5}
\setlength{\tabcolsep}{3pt}

\begin{table*}[h!]
    \centering
    \footnotesize 

\begin{tabular}{l p{0.95\textwidth}}
\toprule
ID & Prompt template \\ \midrule 

0 & 
\begin{minipage}[t]{0.95\textwidth}
\begin{minted}[breaklines,breaksymbolleft=]{text}
Translate 'term_bar' from Bavarian to German.
\end{minted}
\end{minipage} \\ \hline 

1 & 
\begin{minipage}[t]{0.95\textwidth}
\begin{minted}[breaklines,breaksymbolleft=]{text}
Translate 'term_bar' from Bavarian to German. Do not say any other word.
\end{minted}
\end{minipage} \\ \hline

2 & 
\begin{minipage}[t]{0.95\textwidth}
\begin{minted}[breaklines,breaksymbolleft=]{text}
What is the German translation of the Bavarian term 'term_bar'?
\end{minted}
\end{minipage} \\ \hline

3 & 
\begin{minipage}[t]{0.95\textwidth}
\begin{minted}[breaklines,breaksymbolleft=]{text}
What is the German translation of the Bavarian term 'term_bar'? Only output the translated word.
\end{minted}
\end{minipage} \\ \hline

4 & 
\begin{minipage}[t]{0.95\textwidth}
\begin{minted}[breaklines,breaksymbolleft=]{text}
What is the translation of the word 'term_bar' into German?
\end{minted}
\end{minipage} \\ \hline

5 & 
\begin{minipage}[t]{0.95\textwidth}
\begin{minted}[breaklines,breaksymbolleft=]{text}
What is the translation of the word 'term_bar' into German? Only output the translated word. 
\end{minted}
\end{minipage} \\ \hline 

6 & 
\begin{minipage}[t]{0.95\textwidth}
\begin{minted}[breaklines,breaksymbolleft=]{text}
Translate the following word to German 'term_bar'.
\end{minted}
\end{minipage} \\ \hline 

\rowcolor{blanchedalmond} 7 & 
\begin{minipage}[t]{0.95\textwidth}
\begin{minted}[breaklines,breaksymbolleft=]{text}
Translate the following word to German 'term_bar'. Do not say any other word. 
\end{minted}
\end{minipage} \\ \hline 

8 & 
\begin{minipage}[t]{0.95\textwidth}
\begin{minted}[breaklines,breaksymbolleft=]{text}
This is a Bavarian term: 'term_bar'. Only return the normalized Standard German word.
\end{minted}
\end{minipage} \\ \hline 

9 & 
\begin{minipage}[t]{0.95\textwidth}
\begin{minted}[breaklines,breaksymbolleft=]{text}
Translate the Bavarian term: 'term_bar' into High German form. 
\end{minted}
\end{minipage} \\ \hline 

10 & 
\begin{minipage}[t]{0.95\textwidth}
\begin{minted}[breaklines,breaksymbolleft=]{text}
Translate the Bavarian term: 'term_bar' into High German form. Only return the High German form.
\end{minted}
\end{minipage} \\ \hline 

11 & 
\begin{minipage}[t]{0.95\textwidth}
\begin{minted}[breaklines,breaksymbolleft=]{text}
Translate the Bavarian term 'term_bar' into High German form.
\end{minted}
\end{minipage} \\ \hline

\rowcolor{blanchedalmond} 12 & 
\begin{minipage}[t]{0.95\textwidth}
\begin{minted}[breaklines,breaksymbolleft=]{text}
Translate the Bavarian term 'term_bar' into High German form. Only return the High German form.
\end{minted}
\end{minipage} \\ \hline 

13 & 
\begin{minipage}[t]{0.95\textwidth}
\begin{minted}[breaklines,breaksymbolleft=]{text}
Translate the Bavarian term: 'term_bar' into Standard German form. 
\end{minted}
\end{minipage} \\ \hline 

14 & 
\begin{minipage}[t]{0.95\textwidth}
\begin{minted}[breaklines,breaksymbolleft=]{text}
Translate the Bavarian term: 'term_bar' into Standard German form. Only return the Standard German form.
\end{minted}
\end{minipage} \\ \hline 

15 & 
\begin{minipage}[t]{0.95\textwidth}
\begin{minted}[breaklines,breaksymbolleft=]{text}
This is a Bavarian term: 'term_bar'. Only return the normalized Standard German equivalent. Respond with exactly one single word. 
\end{minted}
\end{minipage} \\ \hline 

16 & 
\begin{minipage}[t]{0.95\textwidth}
\begin{minted}[breaklines,breaksymbolleft=]{text}
Translate the word 'term_bar' from Bavarian to German. 
\end{minted}
\end{minipage} \\ \hline

17 & 
\begin{minipage}[t]{0.95\textwidth}
\begin{minted}[breaklines,breaksymbolleft=]{text}
Translate the word 'term_bar' from Bavarian to German. Do not say any other word.
\end{minted}
\end{minipage} \\ \hline

18 & 
\begin{minipage}[t]{0.95\textwidth}
\begin{minted}[breaklines,breaksymbolleft=]{text}
How do you say the Bavarian word 'term_bar' in German? 
\end{minted}
\end{minipage} \\ \hline

19 & 
\begin{minipage}[t]{0.95\textwidth}
\begin{minted}[breaklines,breaksymbolleft=]{text}
How do you say the Bavarian word 'term_bar' in German? Answer only with the translation. 
\end{minted}
\end{minipage} \\ \hline

\rowcolor{blanchedalmond} 20 & 
\begin{minipage}[t]{0.95\textwidth}
\begin{minted}[breaklines,breaksymbolleft=]{text}
Perform translation: convert the Bavarian form 'term_bar' into its Standard German equivalent. Return only the Standard German form, with no additional explanation or formatting.
\end{minted}
\end{minipage} \\ \hline

21 & 
\begin{minipage}[t]{0.95\textwidth}
\begin{minted}[breaklines,breaksymbolleft=]{text}
Perform translation: convert the Bavarian form 'term_bar' into its Standard German equivalent.
\end{minted}
\end{minipage} \\ \hline 

\end{tabular}
   \caption{\textbf{Dialect-to-standard translation} --  \colorbox{blanchedalmond}{Best prompts}: \mistralsmall{} (20), \mistrallarge{} (20), \llamathreesmall{} (20), \llamathreelarge{} (20), \llamafour{} (20), \ayasmall{} (7), \ayalarge{} (20), \gemmasmall{} (12), \gemmalarge{} (20).}
   \label{tab:translation_prompt_templates}
\end{table*}

\begin{table*}[]
\parbox{.45\linewidth}{
\small
\centering
\begin{tabular}{lcc}
\toprule
\textbf{Model} & \textbf{Macro F1} & \textbf{\iferr{}} \\
\midrule

\mistralsmall   & $0.357 \pm 0.129$ & $0.290 \pm 0.451$ \\
\mistrallarge   & $0.368 \pm 0.127$ & $0.163 \pm 0.378$ \\
\llamathreesmall & $0.275 \pm 0.104$ & $0.233 \pm 0.409$ \\
\llamathreelarge & $0.318 \pm 0.092$ & $0.133 \pm 0.387$ \\
\llamafour      & $0.324 \pm 0.090$ & $0.116 \pm 0.317$ \\
\ayasmall       & $0.349 \pm 0.081$ & $0.341 \pm 0.472$ \\
\ayalarge       & $0.337 \pm 0.151$ & $0.354 \pm 0.468$ \\
\gemmasmall     & $0.430 \pm 0.099$ & $0.155 \pm 0.317$ \\
\gemmalarge     & $0.342 \pm 0.138$ & $0.107 \pm 0.281$ \\

\bottomrule
\end{tabular}
\caption{\textbf{Translation candidate judgment} -- Effect of prompt variation across 13 prompts on Macro F1 and \iferr{} ($\pm$ standard deviation) across models, evaluated on the development set (300 items).}
\label{tbl:promptVar_recognition}
}
\hfill
\parbox{.45\linewidth}{
\small
\centering
\begin{tabular}{lll}
\toprule
\textbf{Model} & \textbf{Accuracy} & \textbf{\iferr{}} \\
\midrule
\ayalarge        & $0.184 \pm 0.173$ & $0.528 \pm 0.449$ \\
\ayasmall        & $0.148 \pm 0.148$ & $0.548 \pm 0.447$ \\
\gemmasmall      & $0.193 \pm 0.186$ & $0.457 \pm 0.488$ \\
\gemmalarge      & $0.190 \pm 0.193$ & $0.455 \pm 0.487$ \\
\llamathreesmall & $0.105 \pm 0.104$ & $0.498 \pm 0.489$ \\
\llamathreelarge & $0.210 \pm 0.193$ & $0.461 \pm 0.479$ \\
\llamafour       & $0.217 \pm 0.211$ & $0.528 \pm 0.452$ \\
\mistralsmall    & $0.109 \pm 0.128$ & $0.632 \pm 0.421$ \\
\mistrallarge    & $0.005 \pm 0.016$ & $0.930 \pm 0.199$ \\
\bottomrule
\end{tabular}

\caption{\textbf{Dialect-to-standard translation} -- Effect of prompt variation across 21 prompts on Accuracy and \iferr{} ($\pm$ standard deviation) across models, evaluated on the development set (300 items).}
\label{tbl:promptVar_translation}
}
\end{table*}

\renewcommand{\arraystretch}{1.5}
\setlength{\tabcolsep}{3pt}

\begin{table*}[h!]
    \centering
    \footnotesize 

\begin{tabular}{l p{0.95\textwidth}}
\toprule
ID & Prompt template \\ \midrule

2 & 
\begin{minipage}[t]{0.95\textwidth}
\begin{minted}[breaklines,breaksymbolleft=]{text}
Is the Bavarian term 'term_bar' an exact dialectal variant ('yes'), a dialectal morphological inflection ('inflected'), or not a dialectal variant ('no') of 'term_de' in Standard German? 

Usage example: "####" 

Return only "yes", "inflected", or "no".
\end{minted}
\end{minipage} \\ \hline

3 & 
\begin{minipage}[t]{0.95\textwidth}
\begin{minted}[breaklines,breaksymbolleft=]{text}
Compare 'term_bar' (Bavarian) to 'term_de' (Standard German).

Usage example: "####"

Answer with 'yes' if it's a direct dialectal translation, 'inflected' if it's a dialectal inflected form, or 'no' if neither applies. Do not say any other word.
\end{minted}
\end{minipage} \\ \hline

5 & 
\begin{minipage}[t]{0.95\textwidth}
\begin{minted}[breaklines,breaksymbolleft=]{text}
Task: Is the Bavarian term: "term_bar" a correct dialectal variant of the German term: "term_de"? 

Usage example: "####"

Follow the given annotation guidelines.
Guidelines:
- yes: The candidate is an exact dialectal variation of the Standard German word.
- inflected: The candidate is a morphologically inflected variant of the German word.
- no: None of the two applies.
Return only "yes", "inflected", or "no".
\end{minted}
\end{minipage} \\ \hline 

6 & 
\begin{minipage}[t]{0.95\textwidth}
\begin{minted}[breaklines,breaksymbolleft=]{text}
Classify the Bavarian term 'term_bar' with respect to the Standard German term 'term_de'. 

Usage example: "####"

Return exactly one of the following:
- 'yes' if it is an exact dialectal variant
- 'inflected' if it is a morphologically inflected variant
- 'no' otherwise
\end{minted}
\end{minipage} \\ \hline 

8 & 
\begin{minipage}[t]{0.95\textwidth}
\begin{minted}[breaklines,breaksymbolleft=]{text}
Is the Bavarian term 'term_bar':
- a dialectal translation of the German term 'term_de' → 'yes'
- an inflection of the German term 'term_de' → 'inflected'
- not a variant or inflected variant of the German term 'term_de' → 'no'

Usage example: "####"

Answer only with 'yes', 'inflected', or 'no'.
\end{minted}
\end{minipage} \\ \hline 

\end{tabular}
   \caption{\textbf{Translation candidate judgment (with context)}. We extend for each model the best-performing prompt by including a usage example.}
   \label{tab:recognition_w_context_prompts}
\end{table*}

\renewcommand{\arraystretch}{1.5}
\setlength{\tabcolsep}{3pt}

\begin{table*}[h!]
    \centering
    \footnotesize 

\begin{tabular}{l p{0.95\textwidth}}
\toprule
ID & Prompt template \\ \midrule 

7 & 
\begin{minipage}[t]{0.95\textwidth}
\begin{minted}[breaklines,breaksymbolleft=]{text}
Translate the following word to German 'term_bar'. Usage example: "####". Do not say any other word. 
\end{minted}
\end{minipage} \\ \hline 

12 & 
\begin{minipage}[t]{0.95\textwidth}
\begin{minted}[breaklines,breaksymbolleft=]{text}
Translate the Bavarian term 'term_bar' into High German form. Usage example: "####".  Only return the High German form.
\end{minted}
\end{minipage} \\ \hline

20 & 
\begin{minipage}[t]{0.95\textwidth}
\begin{minted}[breaklines,breaksymbolleft=]{text}
Perform translation: convert the Bavarian form 'term_bar' into its Standard German equivalent.  Usage example: "####". Return only the Standard German form, with no additional explanation or formatting.
\end{minted}
\end{minipage} \\ \hline 

\end{tabular}
   \caption{\textbf{Dialect-to-standard translation (with context)}. We extend for each model the best-performing prompt by including a usage example.}
   \label{tab:translation_w_context_prompts}
\end{table*}

\renewcommand{\arraystretch}{1.5}
\setlength{\tabcolsep}{3pt}

\begin{table*}[h!]
    \centering
    \footnotesize 

\begin{tabular}{l p{0.95\textwidth}}
\toprule
ID & Prompt template \\ \midrule

2 & 
\begin{minipage}[t]{0.95\textwidth}
\begin{minted}[breaklines,breaksymbolleft=]{text}
Ist der bairische Begriff 'term_bar' eine exakte dialektale Variante ('yes'), eine morphologische Beugung ('inflected') oder keine Dialektvariante ('no') von 'term_de' im Hochdeutschen?
\end{minted}
\end{minipage} \\ \hline

3 & 
\begin{minipage}[t]{0.95\textwidth}
\begin{minted}[breaklines,breaksymbolleft=]{text}
Vergleiche 'term_bar' (bairisch) mit 'term_de' (hochdeutsch). Antworten mit 'yes', wenn es sich um eine direkte dialektale Übersetzung handelt, mit 'inflected', wenn es eine gebeugte Form handelt, oder mit 'no', wenn beides nicht zutrifft.
\end{minted}
\end{minipage} \\ \hline

5 & 
\begin{minipage}[t]{0.95\textwidth}
\begin{minted}[breaklines,breaksymbolleft=]{text}
Aufgabe: Ist der bairische Begriff "term_bar" eine korrekte dialektale Variante des hochdeutschen Begriffs "term_de"? Befolge die untenstehenden Annotationsrichtlinien.
Richtlinien:
- yes: Der Begriff ist eine exakte dialektale Entsprechung.
- inflected: Der Begriff ist eine morphologisch gebeugte Variante.
- no: Keines von beiden trifft zu.
Gib nur "yes", "inflected" oder "no" zurück.
\end{minted}
\end{minipage} \\ \hline 

6 & 
\begin{minipage}[t]{0.95\textwidth}
\begin{minted}[breaklines,breaksymbolleft=]{text}
Klassifiziere den bairischen Begriff 'term_bar' im Verhältnis zum hochdeutschen Begriff 'term_de'. Gib genau eines der folgenden Labels zurück:
- 'yes', wenn es eine exakte dialektale Variante ist
- 'inflected', wenn es eine morphologisch gebeugte Form ist
- 'no', andernfalls
\end{minted}
\end{minipage} \\ \hline 

8 & 
\begin{minipage}[t]{0.95\textwidth}
\begin{minted}[breaklines,breaksymbolleft=]{text}
Ist der bairische Begriff 'term_bar':
- eine Dialektübersetzung des deutschen Begriffs 'term_de' → 'yes'
- eine Beugungsform des Deutschen Begriffs 'term_de' → 'inflected'
- keine Variante oder flektierte Variante des deutschen Begriffs 'term_de' → 'no'
Antworte nur mit 'yes', 'inflected', oder 'no'.
\end{minted}
\end{minipage} \\ \hline 

\end{tabular}
   \caption{\textbf{Translation candidate judgment (German prompts)}. We extend for each model the best-performing prompt by including a usage example.}
   \label{tab:recognition_german_prompts}
\end{table*}

\renewcommand{\arraystretch}{1.5}
\setlength{\tabcolsep}{3pt}

\begin{table*}[h!]
    \centering
    \footnotesize 

\begin{tabular}{l p{0.95\textwidth}}
\toprule
ID & Prompt template \\ \midrule 

7 & 
\begin{minipage}[t]{0.95\textwidth}
\begin{minted}[breaklines,breaksymbolleft=]{text}
Übersetze das folgende Wort ins Deutsche 'term_bar'. Gib nur das übersetzte Wort aus.
\end{minted}
\end{minipage} \\ \hline 

12 & 
\begin{minipage}[t]{0.95\textwidth}
\begin{minted}[breaklines,breaksymbolleft=]{text}
Übersetze den bairischen Begriff 'term_bar' ins Hochdeutsche. Gib nur die hochdeutsche Form zurück.
\end{minted}
\end{minipage} \\ \hline

20 & 
\begin{minipage}[t]{0.95\textwidth}
\begin{minted}[breaklines,breaksymbolleft=]{text}
Übersetzung durchführen: konvertiere die bayerische Form 'term_bar' in das hochdeutsche Äquivalent. Gib nur die hochdeutsche Form zurück, ohne weitere Erklärungen oder Formatierungen.
\end{minted}
\end{minipage} \\ \hline 

\end{tabular}
   \caption{\textbf{Dialect-to-standard translation (German prompts)}. We extend for each model the best-performing prompt by including a usage example.}
   \label{tab:translation_german_prompts}
\end{table*}

\setlength{\tabcolsep}{3pt}

\begin{table*}[ht!]
\small
\centering

\begin{tabular}{lrr|rr||rr|rr}
\toprule
& \multicolumn{4}{c||}{\textbf{Context}} & \multicolumn{4}{c}{\textbf{Language}} \\
 & \multicolumn{2}{c|}{\textbf{Judgment}} & \multicolumn{2}{c||}{\textbf{Translation}} 
 & \multicolumn{2}{c|}{\textbf{Judgment}} & \multicolumn{2}{c}{\textbf{Translation}} \\
\textbf{Model} & \textbf{Macro-F1} & \textbf{\iferr{}} & \textbf{Accuracy} & \textbf{\iferr{}}
& \textbf{Macro-F1} & \textbf{\iferr{}} & \textbf{Accuracy} & \textbf{\iferr{}} \\
\midrule
\mistralsmall     &  0.004 &  0.817 & -0.034 &  0.128 & -0.032 &  0.999 &  0.000 & -0.103 \\
\mistrallarge     & -0.016 &  0.366 &  0.183 & -0.005 & -0.174 &  0.351 &  0.023 &  0.011 \\
\llamathreesmall  &  0.028 & -0.003 &  0.147 &  0.047 & -0.028 &  0.022 &  0.012 &  0.000 \\
\llamathreelarge  &  0.136 &  0.000 &  0.130 &  0.032 &  0.122 &  0.000 & -0.015 &  0.002 \\
\llamafour        & -0.272 & -0.433 &  0.110 &  0.033 & -0.255 & -0.437 & -0.021 &  0.001 \\
\ayasmall         & -0.198 &  0.000 & -0.061 &  0.323 & -0.136 &  1.000 & -0.038 &  0.001 \\
\ayalarge         & -0.077 &  0.564 & -0.009 &  0.208 & -0.063 & -0.004 &  0.006 &  0.001 \\
\gemmasmall       &  0.021 &  0.000 &  0.071 &  0.064 & -0.058 &  0.000 & -0.097 &  0.011 \\
\gemmalarge       & -0.036 & -0.001 &  0.095 &  0.046 & -0.184 & -0.001 & -0.023 & -0.001 \\
\bottomrule
\end{tabular}

\caption{Overall changes ($\Delta$) in model performance and instruction-following abilities (\iferr{}) for both tasks. On the left, we report the effect of \textbf{context} (context -- no context), and on the right, the effect of \textbf{prompt language} (German -- English).}
\label{tab:deltas_combined}
\end{table*}

\end{document}